\documentclass[twoside,11pt]{article}

\usepackage{jmlr2e}
\usepackage{amsmath} 

\usepackage{float} 

\usepackage{booktabs}

\ShortHeadings{Improving Decision Tree Predictions via Transformations}{Galili and Meilijson}
\firstpageno{1}

\begin{document}





\title{Splitting matters: how monotone transformation of predictor variables may improve the predictions of decision tree models}

\author{\name Tal Galili \email tal.galili@math.tau.ac.il \\
       \addr Department of Statistics and Operations Research \\
       Tel-Aviv University \\
       Tel Aviv, 69978, Israel
       \AND
       \name Isaac Meilijson \email 
       isaco@math.tau.ac.il \\
       \addr  Department of Statistics and Operations Research \\
       Tel-Aviv University \\
       Tel Aviv, 69978, Israel  
       }

\editor{unknown}

\maketitle

\begin{abstract}


It is widely believed that the prediction accuracy of decision tree models is invariant under any strictly monotone transformation of the individual predictor variables. However, this statement may be false when predicting new observations with values that were not seen in the training-set and are close to the location of the split point of a tree rule. The sensitivity of the prediction error to the split point interpolation is high when the split point of the tree is estimated based on very few observations, reaching 9\% misclassification error when only 10 observations are used for constructing a split, and shrinking to 1\% when relying on 100 observations. This study compares the performance of alternative methods for split point interpolation and concludes that the best choice is taking the mid-point between the two closest points to the split point of the tree. Furthermore, if the (continuous) distribution of the predictor variable is known, then using its probability integral for transforming the variable ("quantile transformation") will reduce the model's interpolation error by up to about a half on average. Accordingly, this study provides guidelines for both developers and users of decision tree models (including bagging and random forest).




\end{abstract}



\section{Introduction}

Algorithms for decision tree learning (DTL) construct a decision tree model (DTM), which uses predictor variables (also known as features or measurements) to match an item to its target value. Decision tree models are very commonly used for predictive modeling in statistics, data mining, and machine learning. DTL is often performed by searching for a series of decision rules which split (partition) the space spanned by the features into disjointed regions of the space. Each partition region of the space is associated with some prediction of the item's target value, which could be categorical or numerical. In these tree structures, leaves include the predicted targets and branches represent conjunctions of features that lead to these predictions. A short survey of decision tree models in general, and the CART methodology in particular, is provided in the appendix at section \ref{app:DTL_intro}.

A classical survey of decision trees’ adaptation is provided by S. K. Murthy \citep{murthy1998automatic}, and a 2013 survey by S. Lomax and S. Vadera concluded that there are (at least) over 50 different algorithms for decision tree learning (including ID3, CHAID, C5.0, oblique trees, etc.) \citep{lomax2013survey}. A drawback of using a single decision tree is that it tends to either give biased predictions or over-fit the data. In recent decades, better alternatives have been presented by extending a single decision tree to an ensemble of decision trees. Leo Breiman is responsible for two celebrated extensions of the CART model - Bagging \citep{breiman1996bagging} and Random Forest \citep{breiman2001random}. Another prominent extension, by Freund and Schapire \citep{freund1995desicion}, is the idea of boosting as implemented in AdaBoost. Comparison of these methods has shown that each model may be better suited to different scenarios \citep{banfield2007comparison}. Further extensions include gradient boosting \citep{friedman2001greedy}.

It is a commonly held paradigm stating that algorithms based on decision tree models are generally invariant under strictly monotone transformations of the individual predictor variables. As a result, scaling and/or more general transformations are not considered an issue to be concerned with (see page 352 in \cite{hastie01statisticallearning}, page 20 in \cite{timofeev2004classification}, or page 181 in \citep{friedman2006recent}). However, the CART book includes a more limited statement (see page 57 in \cite{breiman1984classification}): "In a standard data structure it (the decision tree) is invariant under all monotone transformations of individual \textbf{ordered} variables". Both statements are true if the training data set is supported by all the possible values of the features that would be found when predicting future observations. Specifically, if the training data sets use some ordered explanatory variables that include all the possible values of these features, then using a strictly monotonic transformation should not make a difference in the model's prediction accuracy. 

The aforementioned general statement may not hold when the tree is used for generating an interpolated prediction which is close to the split point location. In general, an interpolated prediction occurs when the model predicts the label of an observation based on a value of the predictor variable that has not been observed in the training data set. This can sometimes happen for discrete predictor variables and will always happen for continuous observations. An interpolated prediction which is close to the split point location is illustrated in the following example. Consider data with only one predictor variable $X$ that can get integer values from 1 to 10, and a deterministic dependent variable $Y$ that gets 0 if $X >= 9$ and 1 otherwise. Suppose that the data used for training the decision tree had only observations with $X$-values 1, 2, and 10 (and $Y$-values 1, 1, and 0). If the model would be used to predict a new observation with $X=1$ it would correctly classify it as 1. But what should the model give when used to interpolate a prediction for an observation that is close to the split point 9, such as $X=8$? There are three methods used in practice by decision tree learners for making such interpolated predictions. The first method, "Sweep Left", classifies any observation above (but not including) 2 as 0 ($X>2 \Rightarrow  Y=0$), and estimates the cutoff quantile ($p$) of the decision tree as $\hat p_{SL}$. Under the second method, "Sweep Right", only observations with the values $X=10$ would be classified as 0 ($X=10 \Rightarrow  Y=0$), with cutoff $\hat p_{SR}$. The third method, with cutoff ${\hat p_{SB}} = \frac{{\hat{p}_{SL} + \hat{p}_{SR}}}{2}$ between the previous two, classifies observations according to the rule $X > \frac{2+10}{2} = 6 \Rightarrow  Y=0 $. In the above example, $\hat p_{SL}$ will correctly classify $X=8$ as 0, while $\hat p_{SR}$ would mistakenly classify the observation as 1. Both of these methods would give the same prediction regardless of whatever monotone transformation would be used on $X$. The third method, $\hat p_{SB}$, would wrongly classify the observation as 0. However, if the variable $X$ is transformed to be $X^2$, then the new rule would classify observations with $X^2 > \frac{2^2+10^2}{2} = 52$ as 0. In this case, $X^2=64$ will be correctly classified as 0. In this example, the monotone transformation $X^2$ influenced (specifically, helped) the interpolated prediction of the decision tree rule for an observations that was close to $X=9$. It is evident that different transformations could either help or damage the prediction under different possible values of $X$, through their influence on the way $\hat p_{SB}$ will make the interpolation prediction. It will be shown that the distribution from which new observations of $X$ arrive could be used as a guide for the best transformation to use in order to minimize the misclassification error of the model on new observations.

All three described methods for split point interpolation prediction are used in practice. A survey of existing R packages \citep{R} reveals a variability in how different decision tree implementations interpolate their split point. The packages tree  \citep{R_tree}, rpart \citep{R_rpart}, oblique.tree \citep{R_oblique_tree}, randomForest \citep{R_randomForest}, and Rborist \citep{R_Rborist} all use $\hat p_{SB}$. The packages C50 \citep{R_C50}, partykit  \citep{hothorn2006unbiased} (i.e., the ctree function \citep{hothorn2015partykit}), RWeka (i.e.: the J48 function which implements Quinlan's C4.5 \citep{hornik2009open}), xgboost \citep{R_xgboost}, and ranger \citep{R_ranger} all use $\hat p_{SL} $. Lastly, the package evtree \citep{R_evtree} uses $\hat p_{SR} $.

As mentioned earlier, the various methods of split point interpolation may disagree on discrete variables and are sure to differ on continuous variables - for new observations that are close to the split point location. There has been an active discussion in the literature on how to treat continuous variables for DTL. Until now, however, research has mostly focused on discretization techniques for dealing with the computational complexity of testing too many potential splits (see \cite{chickering2001efficient, liu2002discretization, kotsiantis2006discretization, fayyad1992handling, kohavi1996error}). As far as we know, no previous work has explored how to make split point interpolation so as to minimize the near-split-point interpolated misclassification error of the model for continuous (or discrete) predictor variables.

To address this issue, we formalize in section (\ref{theory_est_p}) the problem of split point interpolation by introducing the supervised uniform distribution, with random $X$ ($X \sim U(0,1)$) and a deterministic binary $Y$ ($Y = I_{X < p}$), where $p$ is unknown. A variety of estimators for $p$ are proposed, and their statistical properties are investigated. The concluding recommendation will be to use the middle point for interpolation ($\hat p_{SB}$). The effect of near-split-point interpolated prediction error is directly influenced by the sample size used for training the decision tree in each split. For example, if only 10 observations are used for estimating the split point, then $\hat p_{SL}$ could lead to 9\% misclassification error while a model using $\hat p_{SB}$ would reduce it to approximately 4\%. Using 100 observations for training will reduce the misclassification error by a factor of 10 (as compared to 10 observations) to around 0.9\% and 0.4\% respectively. 

Moreover, as discussed earlier in the introduction, the $\hat p_{SB}$ estimator is sensitive to monotone transformations on $X$. Section (\ref{beyond_SU_seq}) illustrates by simulation that if the distribution of a continuous predictor variable is (at least approximately) known, then using its probability integral transformation (termed here ``quantile transformation") will reduce the model's interpolation error by up to half on average (depending on the split point location and the original distribution of $X$). A Bayesian interpretation reveals that using the quantile transformation procedure brings the decision tree closer to taking the median posterior distribution of the split point quantile $p$ (under a uniform prior on $p$). This method is most powerful when the split point is interpolated in an area of the distribution with a monotone but very non-linear density, in which case the median could be far from the mid-point prediction of the two observations in the training set that are used for inducing a node's split point interpolation. The simulation study also explores cases where the cumulative distribution function (CDF) is estimated in various ways, and a case where the distribution of $X$ used for training is different from which future observations are drawn.

Our method is most effective when a decision tree makes a split based on a small training sample, and it would also be more useful for improving algorithms relying on large trees with many nodes, such as bagging or random forests. Our study provides guidelines for designers of decision tree models to use $\hat p_{SB}$ in their implementation (instead of $\hat p_{SL}$ or $\hat p_{SR}$). Also, users of decision trees are advised to use the quantile transformation we propose when there is good parametric knowledge about the distribution of the predictor variables.




\section{Estimating the cutoff quantile $p$ in the supervised uniform distribution} \label{theory_est_p}
\label{sec:SU_p_estimation}

\subsection{The supervised standard uniform distribution}

The problem of split point interpolation can be formulated as follows. Consider a random vector $\underline{X}$ whose entries are $n$ i.i.d. standard uniform random variables $X_i \sim U(0,1) \ , \ i=1 \dots n$, and an unknown parameter $p$ strictly between $0$ and $1$ to be estimated from the following additional data. Each $X_i$ is augmented by the indicator variable $Y_i = I_{X_i < p} $ with value $1$ if $X_i < p$ and $0$ otherwise. Denote by  $\underline{Y}$ the vector with entries $Y_i$. Let $L$ be the largest entry in $\underline{X}$ that is \textbf{L}eft of $p$, and $R$ the smallest entry that is \textbf{R}ight of $p$. Clearly, $Y_i \sim Bernoulli(p)$ and $K=\sum_{i=1}^n Y_i \sim B(n, p)$. Let $X_{(i)}$, the i'th order statistics of $\underline{X}$, where $X_{(1)} = min(\underline{X})$ and $X_{(n)} = max(\underline{X})$, be augmented by $X_{(0)} \equiv 0$ and $X_{(n+1)} \equiv 1$. Since $K$ is the number of observations from $\underline{X}$ that are less than $p$ then $L \equiv X_{(K)} < p < X_{(K+1)} \equiv R$. The density function of a single observation is

\begin{align} \label{XYdensity}
{f_{{X_i},{Y_i}}}\left( {{x},{y};p} \right) = {I_{\left\{ {0<x<p<1,y=1} \right\} \cup \left\{ {0<p<x<1,y=0} \right\}}}
\end{align}
and the joint density of $L$ and $R$ is 
\begin{align} \label{LRdensity}
{f_{L,R}}\left( {l,r} \right) &= I_{\{0=l<p<r<1\}} n{\left( {1 - r} \right)^{n - 1}} \nonumber \\
&+ I_{\{0<l<p<r<1\}} n\left( {n - 1} \right){\left( {l + 1 - r} \right)^{n - 2}} + I_{\{0<l<p<r=1\}} n{l^{n - 1}}
\end{align}

Accordingly, the expectation of a general integrable function $g(L,R)$ is

\begin{align} \label{E_g_LR}
E[g(L,R)] &= n\int_p^1 {g\left( {0,r} \right){{\left( {1 - r} \right)}^{n - 1}}dr} \nonumber \\
&+ n\left( {n - 1} \right)\int_0^p {\int_p^1 {g\left( {l,r} \right){{\left( {l + 1 - r} \right)}^{n - 2}}dr} dl} \nonumber \\
&+ n\int_0^p {g\left( {l,1} \right){l^{n - 1}}dl} 
\end{align}

It will later be shown that the estimation of $p$ should be based on $(L,R)$, and formula (\ref{E_g_LR}) will be instrumental in analytically deriving the bias and variance of various estimators of $p$.

We say that the pair $\left<X_i, Y_i\right>$ is drawn from the \textit{supervised uniform distribution} defined by the endpoints ($a, b$) and the parameter $p \in (a,b)$ ($\left<X_i, Y_i\right> \sim SU(a, b, p)$). This section will focus on the {\em standard} case where $X_i \sim U(0,1)$ and $\left<X_i, Y_i\right> \sim SU(0,1, p)$, which could later be extended to any other distribution.
 
In terms of interpretation, at each node of a decision tree, the DTL is responsible for defining $Y$ based on a split in $X$. The split in $X$ (as defined in that node)  partitions the space, based on optimizing some criterion (such as misclassification, gini, or impurity) with regards to some predicted variable. Regardless of the original distribution of $X$, as long as its cumulative distribution function $F$ is known (or estimated, closely enough, through parametric assumptions about the distribution using labeled and un-labaled observations), then the monotone probability integral transformation (which we will term the \textit{quantile transformation}) could be used to get $F(X) \sim U(0,1)$. Hence, once the DTL defines $Y$ from the response variable, then the couple $\left<F(X), Y\right>$ follows the standard supervised uniform distribution. Furthermore, once a split is made, the following (conditional) split will be on observations that come also from a uniform distribution (since observations from a conditional uniform distribution on an interval are also uniform). Hence, for the following discussion, it is not important how a DTL specifically decides on the partitions as long as the split properly partitions the space into two non-overlapping sets of observations with a different $Y$-behavior ($1$ for when $F(X)<p$ and $0$ otherwise). The results in this study are applicable to any method of decision tree learning, be it a single tree or an ensemble, as long as it is based on a DTL that recursively partitions the space. 
 
Several potential estimators of $p$ shall be introduced in the following sections, and their performance will be explored via the Mean Squared Error (MSE) and Mean Absolute Error (MAE). The latter will be evaluated because, for the purpose of classification trees, care is often taken to minimize the prediction misclassification error of a binary response variable ($Y$). If the cost of an erroneous classification is symmetric (i.e. incorrect classification of $Y=0$ and $Y=1$ are treated equally), then the risk function of $\hat p$ depends on the area under the density function between $p$ and $\hat p$. Here, $\hat p$ is an estimator of the cutoff point parameter $p$ for observations from $U(0,1)$. If $\hat p < p$ then a new observation $obs$ that is between the estimated $\hat p$ and the real $p$ ($\hat p < obs < p$) will be misclassified as 0. The chance this would happen when $X \sim U(0,1)$ (i.e. the expected error rate) is simply $p - \hat p$. Integrating this (and the mirror case of misclassification as $1$) over $\hat p$ is simply the Mean Absolute Error (MAE) function $Error(\hat p, p) = \int_{0}^{1} {|p - x| f_{\hat p}(x)} dx$.

\subsection{The minimal sufficient non-complete statistic for the $SU(0,1, p)$ family of distributions is $(L,R)$}

The goal is to estimate $p$ from $n$ observations coming from the supervised standard uniform distribution. While $p$ can be estimated by the natural unbiased estimator $\hat{p}_Y = \frac{K}{n}$, the statistic $\sum Y_i = K$ ignores information from $\underline{X}$, in which case $\hat{p}_Y$ can be improved. A minimal sufficient statistic that will capture most efficiently all relevant information about the parameter $p$ can be found by studying the likelihood function. Since $\left<X_i, Y_i\right>$ are i.i.d pairs, the likelihood function can be written as
\begin{align} \label{likelihood}
L(p) = \prod\limits_{i = 1}^n {{f_{{X_i},{Y_i}}}\left( {{x_i},{y_i};p} \right)}  = \prod\limits_{i = 1}^n {{I_{\left\{ {{x_i} < p,{y_i} = 1} \right\} \cup \left\{ {{x_i} > p,{y_i} = 0} \right\}}} = {I_{\left\{ {l < p < r} \right\}}}{\rm{ }}}
\end{align} 

The Fisher-Neyman factorization theorem implies that the two-dimensional statistic \linebreak $T(\underline{X}, \underline{Y}) = (L, R)$ is a minimal sufficient statistic for $p$ (see appendix \ref{app:LR_min_suff} for a partial proof). Hence, it is enough to consider estimators of functions of $p$ which are exclusively based on $(L, R)$. 

A natural candidate for estimating $p$ would be the maximum likelihood estimator (MLE). Since the likelihood function is a rectangular function that gets its maximal value $1$ for whichever $p$ satisfies $L \le p \le R$, then the MLE is the closed set $[L,R]$. The MLE is agnostic towards any point estimator for $p$ that is a weighted average of $L$ and $R$. But this does not mean that every such combination of $L$ and $R$ is equally good in estimating $p$. If the estimation of an unknown parameter (such as $p$) relies on a \textit{non-complete} minimal sufficient statistic, it can happen that the MLE would be inefficient (both asymptotically and for finite sample sizes) and that Rao-Blackwell improvements would be non-unique and improvable. See \cite{Galili2016} for a discussion of this behavior. In the case of the supervised uniform distribution the statistic $T=(L, R)$ is a two-dimensional minimal sufficient statistic for estimating a scalar parameter ($p$) and is suspected to be not complete (a partial proof for $n=2$ is given in the appendix, section \ref{app:LR_min_suff}). Since the MLE does not offer a specific point estimator for split point interpolation, alternative point estimators are explored in the next sections.

\subsection{An unbiased estimator for $p$ using $L$ and $R$ (via Rao-Blackwell)}

The Rao-Blackwell theorem \citep{radhakrishna1945information, blackwell1947conditional} offers a procedure (coined ``Rao-Blackwellization'' seemingly by \cite{berkson1955maximum}) for improving a crude unbiased estimator $\hat{\theta}$ of a parameter $\theta$ into a better one (in mean-squared-error or any other convex loss function), by taking the conditional expectation of $\hat{\theta}$
given some sufficient statistic $T$,
i.e., ${\hat \theta_{RB}} = E_\theta [{
		\hat{\theta}|T}]$ (this is a statistic because $T$ is sufficient). 
	
The unbiased estimator $\hat{p}_Y = \frac{K}{n}$  can be improved by Rao-Blackwell based on the minimal sufficient couple $(L, R)$,  
to yield ${{\hat p}_{RB}} = E_p\left[ {{{K \over n}}|L,R} \right]=\frac{1}{n}{E_p}\left[ {K|L,R} \right]$.

If $R=1$, all the $n$ observations are to the left of $p$ and therefore $E_p[K|L,R=1]=n$. If $L=0$, all the $n$ observations are to the right of $p$ and therefore $E_p[K|L=0,R]=0$. If both $L$ and $R$ are strictly between 0 and 1, then it is clear that at least one observation is to the left of $p$ (contributing $1$ to $K$) and at least one is to the right of $p$ (contributing $0$ to $K$). This leaves $n-2$ observations from a Bernoulli distribution with success probability $\frac{L}{{L + 1 - R}}$ to be on the left of $p$. Hence $E_p[K| \ L,R \mbox{ s.t. } 0<L<R<1 ]= 1+0+(n-2)\frac{L}{{L + 1 - R}}$. Combining these results, the Rao-Blackwell unbiased improvement of the estimator $\hat{p}_Y$ is 
\begin{align}
{{\hat p}_{RB}} &= \left\{ {\begin{array}{*{20}{c}}
	0&{L = 0}\\
	{\frac{1}{n} + \frac{{n - 2}}{n}\frac{L}{{L + 1 - R}}}&{0<L<R<1}\\
	1&{R =1 }
	\end{array}} \right. \nonumber \\ 
&= \left[ {\frac{1}{n} + \frac{{n - 2}}{n}\frac{L}{{L + 1 - R}}} \right]{I_{\left\{ {0 < L<R < 1} \right\}}} + {I_{\left\{ {R = 1} \right\}}}
\label{p_RB}
\end{align}

If $L>0$ and $R<1$ happen to be very close to each other then the estimator for $p$ from eq. (\ref{p_RB}) is seen to obtain the value ${\hat p}_{RB} \approx L+{{1-2L} \over n}$, showing that ${{\hat p}_{RB}}$ may be outside the feasibility interval $(L,R)$ if the latter is short enough.




The variance of ${\hat p}_{RB}$ coincides with its $MSE$ and, for $n \ne 3$, is

\begin{align} \label{RB_variance}
MSE_p^{(n \ne 3)}\left[ \hat{p}_{RB} \right] = V_p^{(n \ne 3)}\left[ {{{\hat p}_{RB}}} \right] 
&= \frac{1}{{(n - 3)n}}\left[ {\frac{{\left( {n - 1} \right)}}{n}\left[ {1 - \left( {{p^n} + {{(1 - p)}^n}} \right)} \right] - 2p\left( {1 - p} \right)} \right] \nonumber \\
&\approx {{1-2p(1-p)} \over n^2}
\end{align}
Specifically, for $n=2$ the variance is
$V_p^{(n=2)}\left[ {{{\hat p}_{RB}}} \right] 
= \frac{1}{2}(1 - p)p
$. This is as it should be, since for this case ${\hat p}_{RB}=\hat{p}={K \over 2}$. More details are provided in the appendix in section \ref{app:p_RB_extra_equations} (specifically see eq. (\ref{RB_variance_extended}) and eq. (\ref{RB_variance_3})).

Since $(L, R)$ is not complete, the Lehmann-Scheff\'{e} theorem \citep{lehmann1950completeness, lehmann1955completeness2} does not hold, and this estimator may or may not have minimal variance among the unbiased estimators of $p$. In fact, there may not exist an unbiased estimator of $p$ with uniformly minimal variance. As will be seen in the next section, there exist estimators of $p$ with very small bias but with $MSE$ that is noticeably smaller than that of ${\hat p}_{RB}$ for all values of $p$ (other than $p=0$ and $p=1$). Such is the case for ${\hat p}_{B}={{L+R} \over 2}$, with $MSE$ of order of magnitude ${1 \over {2n^2}}$ (constant in $p$), which coincides with the RHS of eq. (\ref{RB_variance}) for $p={1 \over 2}$ but exceeds it otherwise, although never reaching as much as twice. In order to produce the MAE for ${\hat p}_{RB}$ curves in Figure \ref{fig:MAE_curves} they were calculated using Monte Carlo methods, since we could not derive them analytically.


\subsection{Estimating $p$ using $L$ and $R$ separately}

In this section $\hat{p}_L=L$ and $\hat{p}_R=R$ are considered individually for estimating $p$, and their bias, variance, MSE, and MAE terms are evaluated. The following calculations suggest that there may not be a way to create unbiased estimators when solely relying on $L$ or $R$.

The expectation of $\hat{p}_L$ can be evaluated applying $g\left( {L,R} \right) = L$ to eq. (\ref{E_g_LR}) which leads to

\begin{align}
E_p\left[ \hat{p}_L \right] = E_p\left[ L \right]
&= p - \frac{{1 - {{\left( {1 - p} \right)}^{n + 1}}}}{{n + 1}} \label{EL}
\end{align}

Combined with $E\left[ {{L_S}^2} \right]$ (see eq. (\ref{E_L2}) in section \ref{sec:estimating_p_using_L}), the variance $V_p(\hat{p}_L)$ is

\begin{align}
V_p\left[ \hat{p}_L \right] 
&= \frac{{\frac{n}{{\left( {n + 2} \right)}} - 2\left( {np - \frac{1}{{\left( {n + 2} \right)}}\left( {1 - p} \right)} \right){{(1 - p)}^{n + 1}} - {{(1 - p)}^{2(n + 1)}}}}{{{{(n + 1)}^2}}}
 \label{VL}
\end{align}
leading to
\begin{align}
MSE_p\left[ \hat{p}_L \right] 
&= \frac{{2\left( {1 - \left( {p\left( {n + 1} \right) + 1} \right){{(1 - p)}^{n + 1}}} \right)}}{{(n + 1)(n + 2)}} \approx {2 \over {n^2}} \label{MSEL}
\end{align}
which (for large $n$) is four times the MSE of the Bayes estimator and at least twice the MSE of the RB estimator. It should be observed in eq. (\ref{EL}) that the bias of $L$, as an estimator of $p$, is in the same order of magnitude as its standard error. 

Also, the MAE is 

%
%

\begin{align}
MAE_p\left[ \hat{p}_L \right] = E\left[ {\left| {\hat{p}_L - p} \right|} \right] = p - E\left[ L \right]	= \frac{{1 - {{(1 - p)}^{n + 1}}}}{{n + 1}} \approx \frac{1}{n}
\end{align}

For the sake of completeness, $E_p(\hat{p}_R)$ and $V_p(\hat{p}_R)$ are given by
\begin{align}
E_p\left[ \hat{p}_R \right] = E_p\left[ R \right] 
&= p + \frac{{1 - {p^{n + 1}}}}{{ {n + 1} }} \label{ER}
\end{align}
and
\begin{align}
V_p\left[ \hat{p}_R \right] 
&= \frac{{\frac{n}{{\left( {n + 2} \right)}} - 2\left( {n(1 - p) - \frac{1}{{\left( {n + 2} \right)}}p} \right){p^{n + 1}} - {p^{2(n + 1)}}}}{{{{(n + 1)}^2}}}
\label{VR}
\end{align}
leading to
\begin{align}
MSE_p\left[ \hat{p}_R \right] 
&= \frac{{2\left( {1 - ((1 - p)\left( {n + 1} \right) + 1){p^{n + 1}}} \right)}}{{(n + 1)(n + 2)}} \approx {2 \over {n^2}} \label{MSER}
\end{align}

And also, by using eq. (\ref{ER}), the MAE is


%

\begin{align}
MA{E_p}\left[ \hat{p}_R \right] = E\left[ {\left| {\hat{p}_R - p} \right|} \right]= E\left[ { {R } } \right] - p = \frac{{1 - {p^{n + 1}}}}{{n + 1}} \approx \frac{1}{n}
\end{align}

As expected, the variance, MSE, and MAE of $\hat{p}_R$ and $\hat{p}_L$ are obtained from each other by exchanging the roles of $p$ and $1-p$.

\subsection{An optimal Bayes rule for estimating $p$ using $L$ and $R$ together}

From an ad-hoc Bayesian perspective, consider as prior on $p$ the uniform distribution $U(0,1)$. Since the likelihood is constant wherever positive, the posterior distribution of $p$ is $U(L,R)$. This makes the Bayes estimator (under square loss) to be the posterior expectation
\begin{align} 
{{\hat p}_{B}} = E[p|\underline{X},\underline{Y}] = \frac{{L + R}}{2} 
\end{align}
As a proper Bayes rule, this estimator is biased but admissible. From eq. (\ref{EL}) and (\ref{ER}), its expectation (as a function of $p$) is  

\begin{align} \label{E_p_B}
E_p\left[ {\frac{{L + R}}{2}} \right] 
 = \frac{1}{2}E_p\left[ L \right] + \frac{1}{2}E_p\left[ R \right] 
&= p + \frac{{{{\left( {1 - p} \right)}^{n + 1}} - {p^{n + 1}}}}{{2\left( {n + 1} \right)}}
\end{align}

Its variance is

\begin{align}
V_p\left[ {\frac{{L + R}}{2}} \right] 
&= \frac{{1 - (n + 2)p\left( {1 - p} \right)\left( {{p^n} + {{(1 - p)}^n}} \right)}}{{2(n + 1)(n + 2)}} - \frac{{{{\left( {{{(1 - p)}^{n + 1}} - {p^{n + 1}}} \right)}^2}}}{{4{{(n + 1)}^2}}} \approx {1 \over {2n^2}} 
\label{VarBayes}
\end{align}

And combining the two leads to the following MSE 
\begin{align} \label{MSE_p_B}
MSE_p\left[ {\frac{{L + R}}{2}} \right]
&= \frac{{1 - (n + 2)(1 - p)p\left( {{p^n} + {{(1 - p)}^n}} \right)}}{{2(n + 1)(n + 2)}} \approx {1 \over {2n^2}} \nonumber
\end{align}
Variance and MSE are symmetric around $p=0.5$, as expected.

The Mean Absolute Error function is minimized by taking the median of the posterior (uniform) distribution, which, yet again, is ${{\hat p}_{B}} = \frac{{L + R}}{2}$, leading to the following MAE


\begin{align}
MA{E_p}\left[ {\frac{{L + R}}{2}} \right] &= E\left[ {\left| {\frac{{L + R}}{2} - p} \right|} \right] 
\nonumber\\
&= \frac{{1 - \left( {{p^{n + 1}} + {{\left( {1 - p} \right)}^{n + 1}}} \right) + {{\left( {\left| {p - \left( {1 - p} \right)} \right|} \right)}^{n + 1}}}}{{2(n + 1)}} 
\approx \frac{1}{2n}
\end{align}

\subsection{Sweeping estimators for $p$}

As mentioned in the introduction, a modification of the Bayes estimator that is commonly applied as a split rule lets $\hat p$ be taken as $0$ rather than $R \over 2$ if $L=0$, and as $1$ rather than $\frac{L+1}{2}$ if $R=1$. This modified estimator will be called Swept Bayes ($\hat p_{SB}$), as it is obtained from $\hat p_{B}$ by ``sweeping'' mass to the endpoints. The motivation for the terminology comes from a similar concept “b\'{a}l\`{a}yage” or ``sweeping'' used in martingales \citep{picco2012classical}. Similarly, an estimator $\hat{p}_{SL}$ alternative to $\hat{p}_L$ will get value $1$ (instead of $L$) if $R=1$ and the corresponding $\hat{p}_{SR}$ gets $0$ (instead of $R$) if $L=0$.

\begin{align} \label{sweeping_estimators}
\hat{p}_{SL} &= {I_{\left\{ {R < 1} \right\}}}L + {I_{\left\{ {R = 1} \right\}}} \\
\hat{p}_{SR} &= {I_{\left\{ {0 < L} \right\}}}R\\
{\hat p_{SB}} &= \frac{{\hat{p}_{SL} + \hat{p}_{SR}}}{2} = \frac{{L + R}}{2}{I_{\left\{ {0 < L} \right\}}}{I_{\left\{ {R < 1} \right\}}} + {I_{\left\{ {R = 1} \right\}}} 
\end{align}

The expectation of $\hat{p}_{SL}$ and $\hat{p}_{SR}$ resembles that of $\hat{p}_{L}$ and $\hat{p}_{R}$, but with an added term to the bias.

\begin{align}
{E_p}\left[ \hat{p}_{SL} \right] &= p - \frac{{1 - {{\left( {1 - p} \right)}^{n + 1}}}}{{n + 1}} + {p^n}\left( {1 - \frac{n}{{n + 1}}p} \right) \\
{E_p}\left[ \hat{p}_{SR} \right] &= p + \frac{{1 - {p^{n + 1}}}}{{\left( {n + 1} \right)}} - \frac{{\left( {np + 1} \right){{\left( {1 - p} \right)}^n}}}{{\left( {n + 1} \right)}} \\ 
E\left[ {{{\hat p}_{SB}}} \right] &= p + \frac{1}{2}\left( {{p^n}\left( {1 - p} \right) - p{{\left( {1 - p} \right)}^n}} \right) \label{E_R_SB}
\end{align}

The MSE of ${\hat p}_{SB}$ is:


\begin{align} \label{MSE_R_SB}
	MSE\left[ {\frac{{{\hat{p}_{SL}} + {\hat{p}_{SR}}}}{2}} \right] = \frac{1}{{4(n + 1)(n + 2)}}\left[ \begin{array}{l}
	3n\left( {n + 3} \right)\left( \begin{array}{l}
	{\left( {1 - p} \right)^2}{p^n}\\
	+ {p^2}{(1 - p)^n}
	\end{array} \right)\\
	+ 2\left( \begin{array}{l}
	3\left( {{{\left( {1 - p} \right)}^n} + {p^n}} \right) + 1\\
	- 2\left( \begin{array}{l}
	\left( {2 + p} \right){\left( {1 - p} \right)^{n + 1}}\\
	+ \left( {2 + \left( {1 - p} \right)} \right){p^{n + 1}}
	\end{array} \right)
	\end{array} \right)
	\end{array} \right]
\end{align}

Similar to eq. (\ref{E_p_B}) and eq. (\ref{MSE_p_B}), the bias and MSE of ${\hat p_{SB}}$, given in eq. (\ref{E_R_SB}) and eq. (\ref{MSE_R_SB}), reveals a symmetry of the values around $p=0.5$.


\subsection{Comparison of estimators for $p$}


Since different software packages choose different methods for split point interpolation, the question is which of the methods is best among ${{L+R} \over 2}$, $L$, $R$, their swept versions, or the Rao-Blackwell estimator? From the previous section, when the sample size ($n$) is large, the Root Mean Square Error (RMSE) of the different estimators is as follows

\begin{align} \label{RMSE_summary1}
RMS{E_p}\left( {{{\hat p}_{RB}}} \right) &\approx \frac{\sqrt{1 - 2p(1 - p)}}{{n}}  \\ \label{RMSE_summary2}
RMS{E_p}\left( \hat{p}_{L} \right) = RMS{E_p}\left( \hat{p}_{R} \right) &\approx \frac{\sqrt{2}}{{n}}  \\ \label{RMSE_summary3}
RMS{E_p}\left( \hat{p}_{B} \right) &\approx \frac{1}{{\sqrt{2}{n}}}
\end{align}

\begin{figure}[h]
\centering
\includegraphics[width=0.99\linewidth]{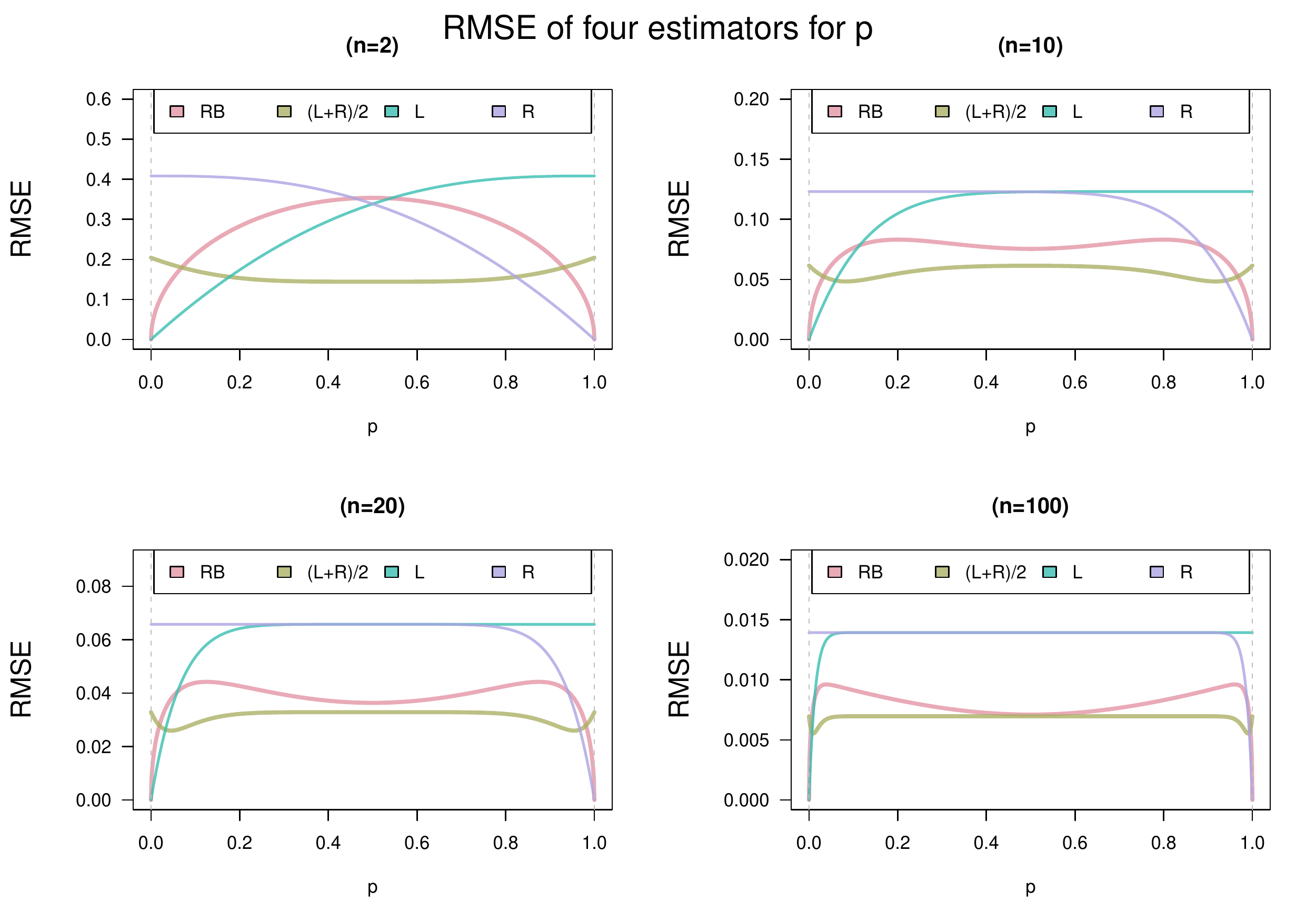}
\caption{The RMSE of the four estimators, across $p$, for different sample sizes ($n=2,10,20,100$).}
\label{fig:RMSE_curves}
\end{figure}

From equations (\ref{RMSE_summary1}, \ref{RMSE_summary2}, \ref{RMSE_summary3}) it is clear that all four estimators are $1 \over n$-consistent in both RMSE and in probability, and the same holds true for the sweep versions of the estimators. For reasonably large $n$ and $p \approx 0.5$,
$RMS{E_p}\left( {{{\hat p}_{RB}}} \right) \approx RMS{E_p}\left( \hat{p}_{B} \right) \approx \frac{1}{2}RMS{E_p}\left( \hat{p}_{L} \right) \approx \frac{1}{2}RMS{E_p}\left( \hat{p}_{R} \right)$, but $RMS{E_p}\left( {{{\hat p}_{RB}}} \right)$ increases as $p$ deviates from 0.5. Figure \ref{fig:RMSE_curves} illustrates the RMSE behavior of the four estimators for various values of $p$ under sample sizes $n=2,10,20,100$.

Figure \ref{fig:RMSE_curves_sweep} (in the appendix, section \ref{sec:performance_of_p_est}) incorporates the sweep estimators as dashed lines. The unbiased Rao-Blackwell and the three swept versions are calibrated so as to decide that there is only one class in the population whenever this is the case in the sample, whatever the sample size. This property makes these estimators have vanishing MSE at extreme values of $p$. But this very same property makes these estimators pay a relatively high price in terms of RMSE at a range of nearly extreme values of $p$. The Bayes rule has stable, nearly constant RMSE, paying a price for ignoring the one-class scenario only at extreme $p$. Figure \ref{fig:RMSE_curves_sweep} shows (for sample sizes $n>=10$) that the one-sided estimators of $p$ (relying on only L or R) have roughly twice the RMSE of the symmetric-type estimators and should thus be avoided. Furthermore, the results in Figure \ref{fig:RMSE_curves_sweep} reveal that the Swept Bayes estimator ($\hat{p}_{SB}$) fully dominates the Rao-Blackwell estimator ($\hat{p}_{RB}$) in RMSE, for all possible values of $p$. For the sake of brevity, the rest of this paper will ignore the sweep estimators, since assertions for comparing $\hat p_B$ to $\hat p_L$, $\hat p_R$ and $\hat p_{RB}$ will be similar when discussing their swept versions.
 
\begin{figure}[h]
	\centering
	\includegraphics[width=0.99\linewidth]{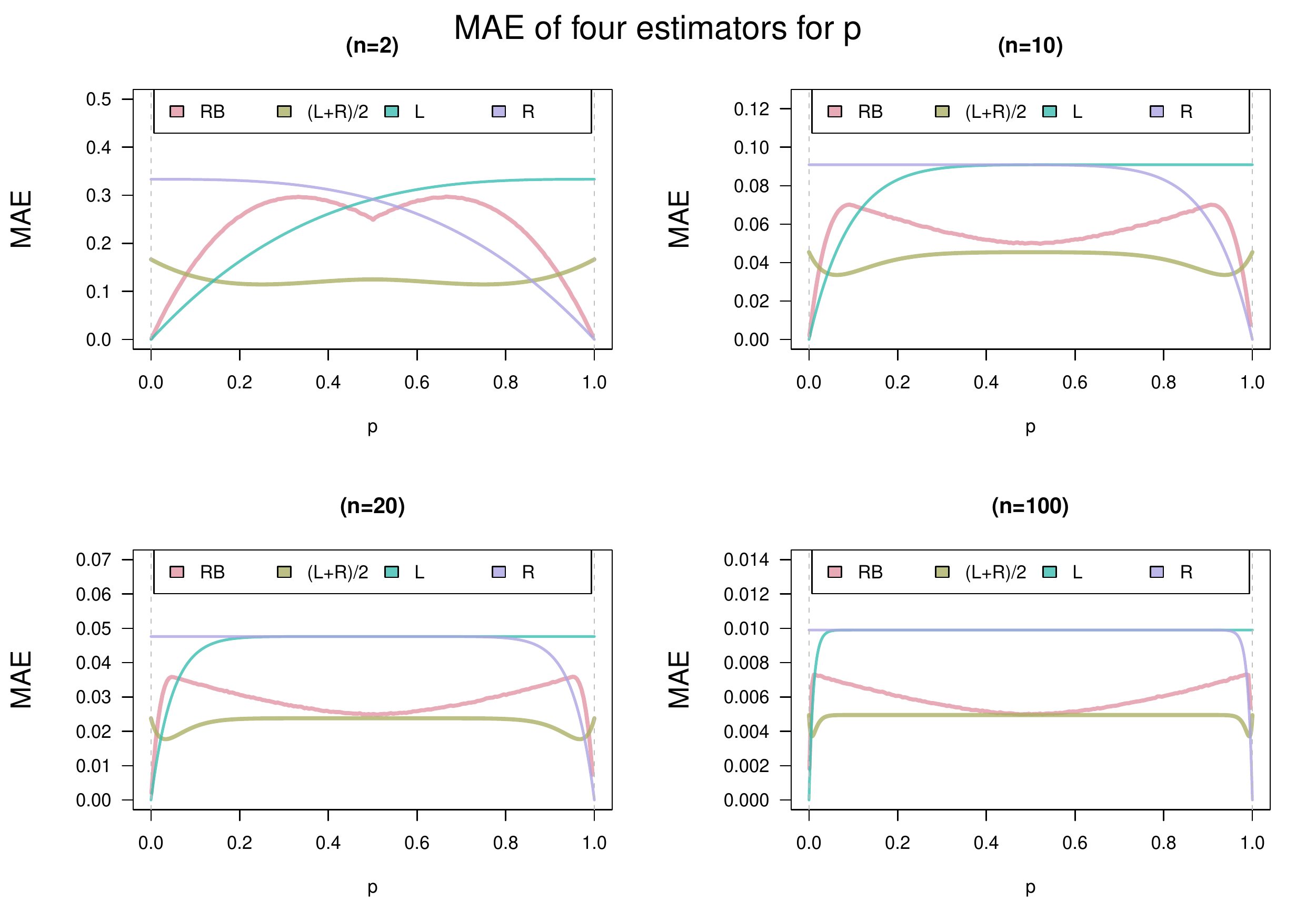}
	\caption{The MAE of the four estimators, across $p$, for different sample sizes ($n=2,10,20,100$).}
	\label{fig:MAE_curves}
\end{figure}

Repeating this analysis on the MAE reveals a similar behavior as shown for the MSE - $MA{E_p}\left( \hat{p}_{B} \right) \approx \frac{1}{2}MA{E_p}\left( \hat{p}_{L} \right) \approx \frac{1}{2}MA{E_p}\left( \hat{p}_{R} \right)$. While the MAE for ${{{\hat p}_{RB}}}$ was not derived analytically, it can still be compared to the other estimators of $p$ as displayed in figure \ref{fig:MAE_curves}, where the curve was estimated through simulation ($10^5$ simulations per point). This figure shows a similar pattern as was seen for RMSE in Figure \ref{fig:RMSE_curves}.

In the context of decision tree models for predicting a binary outcome with an observed predictor variable from $U(0,1)$ (i.e. supervised uniform), the MAE (Mean Absolute Error) is the expected misclassification error of the model. When comparing the four estimators based on MAE, it is clear that the Bayes estimator is better (i.e. has lower MAE) than the Rao-Blackwell estimator for most values of $p$, except for $p$ close to 0 or 1. The larger the sample size, the closer $p$ needs to be to the edges of the support in order for Rao-Blackwell to improve on Bayes, and if $p=0.5$ then the two estimators' performance coincide. The $\hat{p}_{L}$ and $\hat{p}_{R}$ estimators produce double the MAE than the Bayes estimator, unless $p$ is near 0 or 1. These findings lead to the conclusion that unless $p$ is known to lie close to 0 or 1, the best estimator to use is $\hat p_{B} = {{L+R} \over 2}$ (or $\hat p_{SB}$, which is simpler to implement).


\section{Beyond the supervised uniform distribution} \label{beyond_SU_seq}

\subsection{Transforming predictors to the supervised uniform distribution}

Previous sections demonstrated the superiority of $\hat p_{B} = {{L+R} \over 2}$ for interpolating the split point when $X \sim U(0,1)$. Under a uniform prior over the split point $p$, this estimator is Bayes-optimal for squared error (MSE) as well as for absolute difference error functions (MAE) since it is the mean and median of the posterior distribution. It gives the best point-wise result in MSE and MAE over all non-extreme values of $p$. 

Using $\hat p_{B}$ would also be the best solution for any $X \sim U(a,b)$ since it would still be the median (and mean) of the posterior distribution. However, this would no longer hold when $X$ comes from a non-uniform cumulative distribution $F$. If $F$ is known, it is possible to transform $X$ back into the uniform distribution using $F(X) \sim U(0,1)$ (termed quantile transformation) and thus return to the supervised uniform distribution problem ( $\left<F(X_i), Y_i\right> \sim SU(0,1, p)$ ). The proposed algorithm is to first use the quantile transformation on the predictor variables before training the decision tree ($F(X_{train})$), and then apply the same quantile transformation on the new observations ($F(X_{test})$) before predicting their outcome using the trained decision tree model. Since the quantile transformation is monotone, the performance when predicting the training data using the trained decision tree model would be invariant to whether the transformation was used or not. However, for the predictions of new observations (for most possible cases of $p$), using the quantile transformation is expected to improve the interpolated misclassification error of the model (MAE).

Let $L_X$ (and $R_X$ respectively) be the maximal (minimal) observation Left (Right) of the $p$ quantile in the $X$-scale. In our proposed algorithm, $\hat p_{B}$ can be expressed in terms of the statistic ($L_X,R_X$) as $\hat p_{B} = {{F(L_X)+F(R_X)} \over 2} = {{L+R} \over 2} $. If F is not known, one could still use ${{L_X+R_X} \over 2}$ as a split-point on the $X$-scale. This will typically be somewhat distant from $\hat p_{B}$ as it would be like estimating $p$ in the uniform-scale as

\begin{align}
\hat p_X = F \left( {{L_X+R_X} \over 2} \right)
\end{align}

For general $F$, $\hat p_X$ is likely to be distant from the median of the posterior distribution, and therefore give suboptimal misclassification error when predicting new observations (MAE). This shall be explored in the following sections via simulations. 

Whether using $X$ or $F(X)$, using only $F(L_X)$ or $F(R_X)$ for estimating $p$ would be the same as using $L$ or $R$, which would give (approximately) double the MAE (for most values of $p$) and should therefore be avoided.

\subsection{The benefit of transforming the predictor variable in various distributions}

In order to investigate the benefit of knowing $F$, allowing to use $\hat p_{B}$ (or $\hat p_{SB}$) instead of $\hat p_X$, several simulations were conducted for various known distributions. Each simulation checked a range of possible $p$ locations and sample sizes - measuring the Mean Absolute Error (MAE, or misclassification error - since this is measured on the quantile scale) for each of the two estimators ($\hat p_{B}$ vs $\hat p_X$).

\begin{figure}[h]
	\centering
	\includegraphics[width=0.99\linewidth]{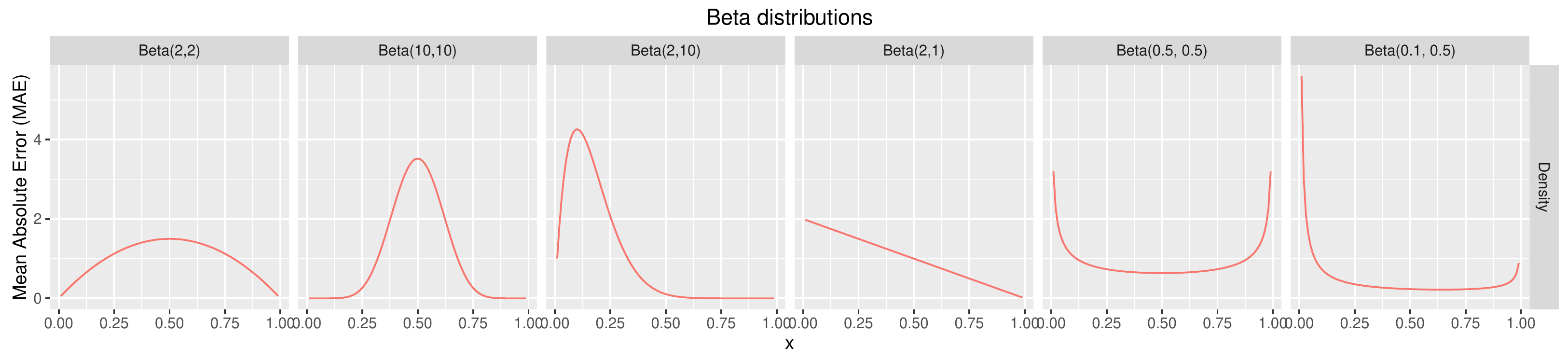}
	\caption{The densities of six parameter combinations for the Beta distribution.}
	\label{fig:Beta_dist_densities}
\end{figure}

Each simulation measured, on the quantile scale, the misclassification error (MAE) for a range of scenarios with sample sizes 2, 10, 20, and 100, on eight Beta distributions - as depicted in Figure \ref{fig:Beta_dist_densities}. The split point $p$ is always taken to be all multiples of 0.01 in $(0,1)$. Each iteration in the simulation draws $n$ observations from $U(0,1)$ and calculates the absolute difference between $p$ and $\hat p_{B}$ (already depicted in Figure \ref{fig:MAE_curves}), as well as between $p$ and $\hat p_X$. This was repeated $10^5$ times and averaged to produce the (MAE) lines in the figures.

\begin{figure}[h]
	\centering
	\includegraphics[width=0.99\linewidth]{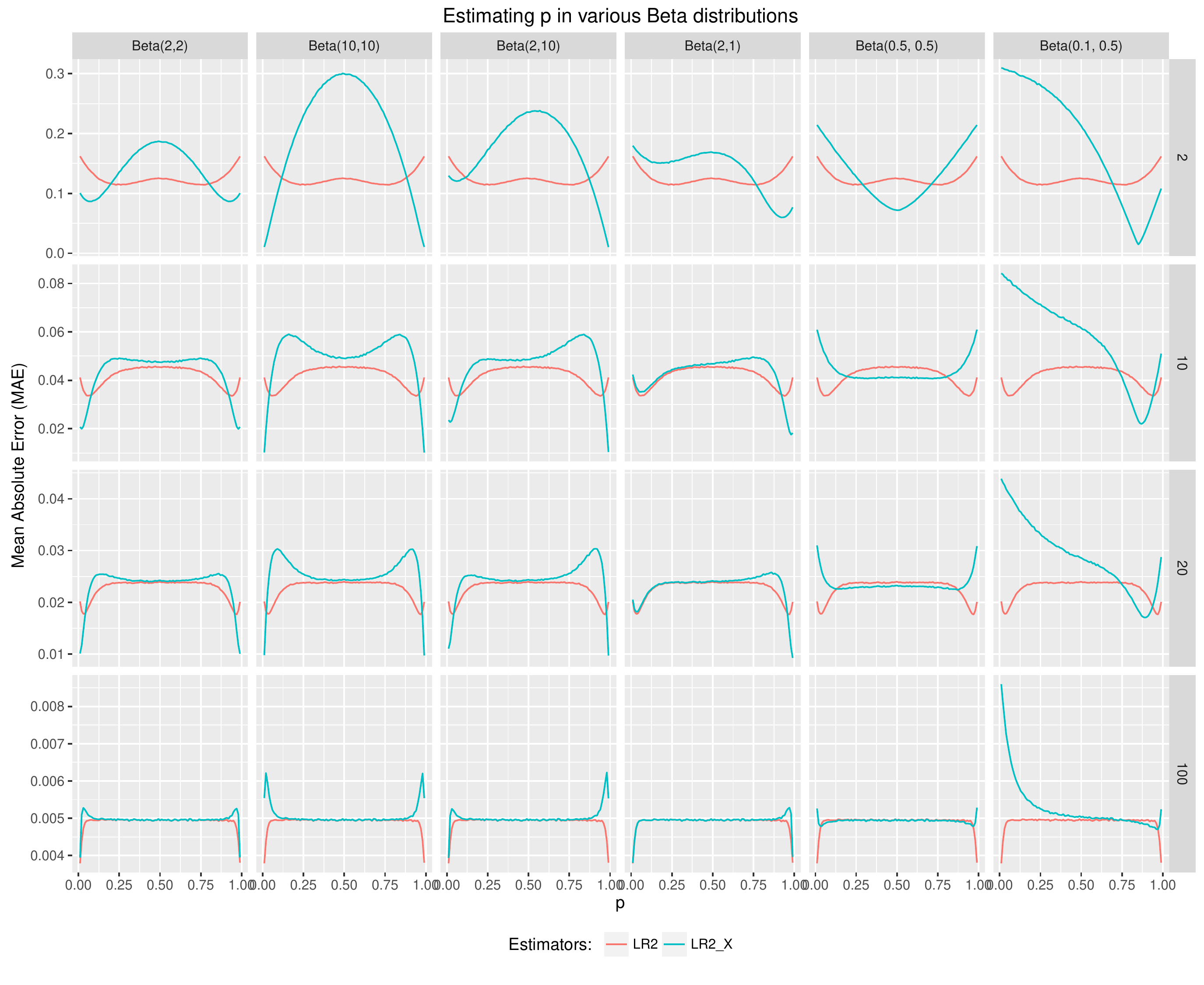}
	\caption{The misclassification error (Mean Absolute Error) of a model estimating $p$ using the original data or the quantile transformed data, for a range of $p$ positions and sample sizes. In the figure $LR2=\hat p_{B}$ and $LR2_X=\hat p_X$.}
	\label{fig:Beta_simu_LR2_LR2X}
\end{figure}

As can be seen from Figure \ref{fig:Beta_simu_LR2_LR2X}, the advantage of using $\hat p_{B}$ over $\hat p_X$ is highly dependent on the shape of the distribution, the location of the split ($p$), and the sample size. Although $\hat p_X$ can have a lower MAE than $\hat p_{B}$ at some $p$-ranges, $\hat p_{B}$ generally performs better and can never be fully dominated by the performance of $\hat p_X$ (as Bayes estimators are admissible). Using $\hat p_{B}$ instead of $\hat p_X$ offers a gain that increases the steeper the density is near the split point ($p$). For example, notice the hump for $Beta(2,10)$ and $n=10$ near $p=0.8$ (second row from the top, and third column from the left, in Figure \ref{fig:Beta_simu_LR2_LR2X}). $\hat p_{B}$ is expected to make approximately 4\% misclassification error, as compared to approximately 6\% if using $\hat p_{X}$. Figure \ref{fig:Beta_dist_densities} (third column from the left) displays a steep decline of this density near its 80\% quantile 0.248. In such a case $\hat p_{X}$ acts similarly to $\hat p_{R}$ in the quantile scale, while $\hat p_{B}$ (generated by the quantile transformation $F(X)$) gives better results, invariant in $F$, by estimating $p$ as the median of the posterior distribution (which, as was shown in the previous section, is the optimal flat-prior Bayes solution for minimizing the MAE).

\begin{figure}
	\centering
	\includegraphics[width=0.99\linewidth]{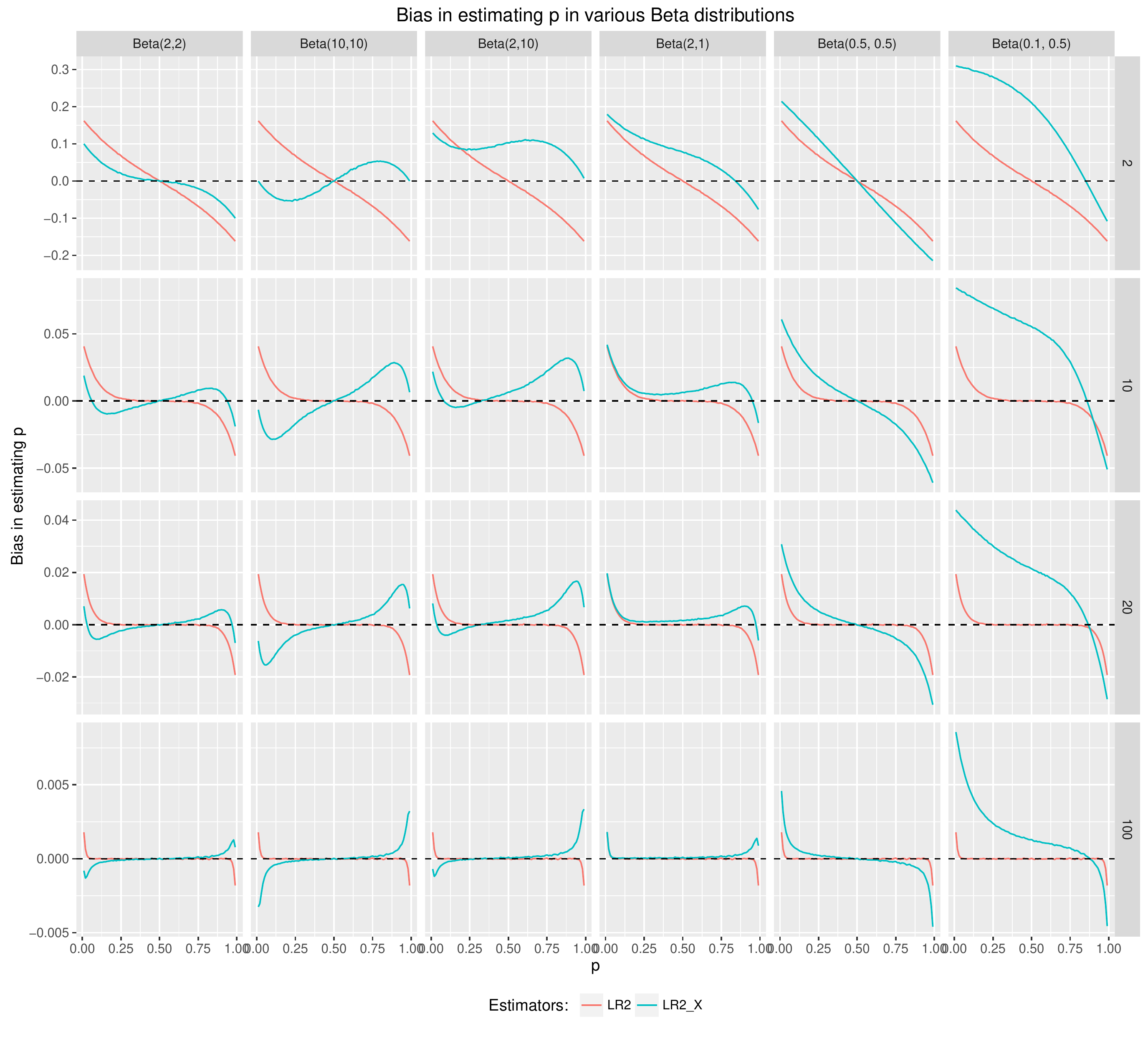}
	\caption{The bias of a model estimating $p$ using the original data or the quantile transformed data, for a range of $p$ positions and sample sizes}
	\label{fig:Beta_simu_LR2_LR2X_bias}
\end{figure}

The Bayes estimator is less precise when the true value of $p$ is near the edges of the support (near 0 or 1). From Figure \ref{fig:Beta_simu_LR2_LR2X_bias} it seems that the raw $X$-scale estimator (not quantile-transformed) is less biased near the edges of $p$, thus helping the estimator gain more precision in these areas over the Bayes estimator. Similar simulations were conducted for other commonly used distributions (Cauchy, Standard Normal, Double Exponential, Chi-squared with $df=1$, Standard Exponential, Log-Normal, and a mixture of two normals), with the results presented in Figure \ref{fig:Common_dist_simu_LR2_LR2X}. From both figures (\ref{fig:Common_dist_simu_LR2_LR2X}) and (\ref{fig:Beta_simu_LR2_LR2X}) it is clear that a larger sample size both reduces the MAE (as expected) and also changes the locations in which $\hat p_{B}$ is better than $\hat p_X$. For example, in unimodal and symmetric distributions (such as Beta(2,2), Beta(10,10), Cauchy, Standard Normal, and the Double Exponential), the larger the sample size gets the more the MAE of both estimators becomes similar for values of $p$ close to 0.5, while for values of $p$ closer to 0 and 1, $\hat p_{B}$ demonstrates better MAE than $\hat p_X$. We note that in no case does one estimator completely dominate the other, and also that only for $n=2$ does one estimator perform twice as good than the other for $p$ near 0.5 (i.e. $\hat p_X$ in this case performs similarly to using only $L$ or $R$).

\begin{figure}[h]
	\centering
	\includegraphics[width=0.99\linewidth]{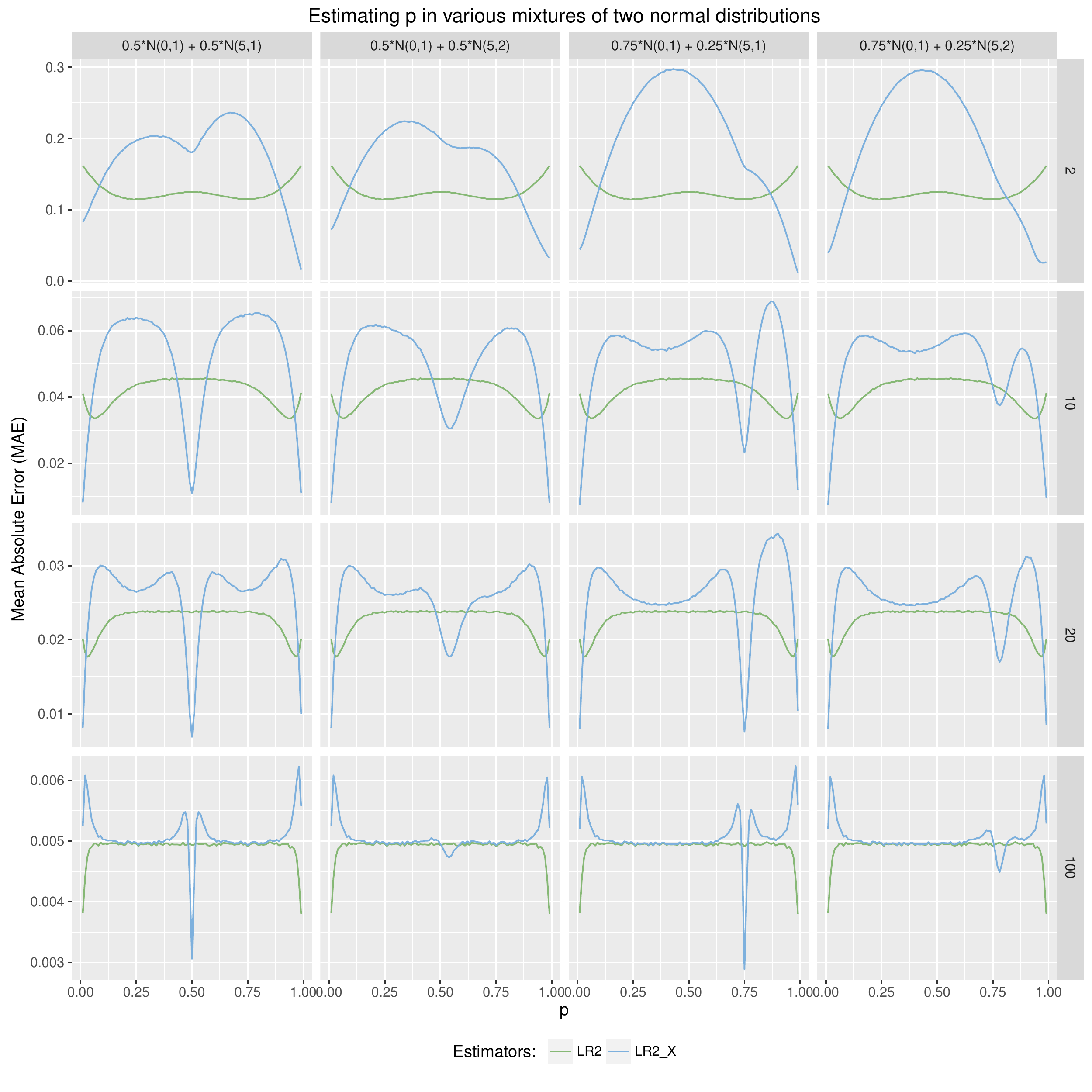}
	\caption{The misclassification error (Mean Absolute Error) of a model estimating $p$ using the original data or the quantile transformed data, for a range of $p$ positions, sample sizes, for four combination of two mixed densities of the standard normal distribution in chances of $0.5$ and $0.75$ with another distribution (chances of $0.5$ and $0.25$) with $\mu=5$ and $\sigma=1$ or $\sigma=2$. In the figure $LR2=\hat p_{B}$ and $LR2_X=\hat p_X$.}
	\label{fig:normmix_dist_simu_LR2_LR2X}
\end{figure}

\begin{figure}[h]
	\centering
	\includegraphics[width=0.99\linewidth]{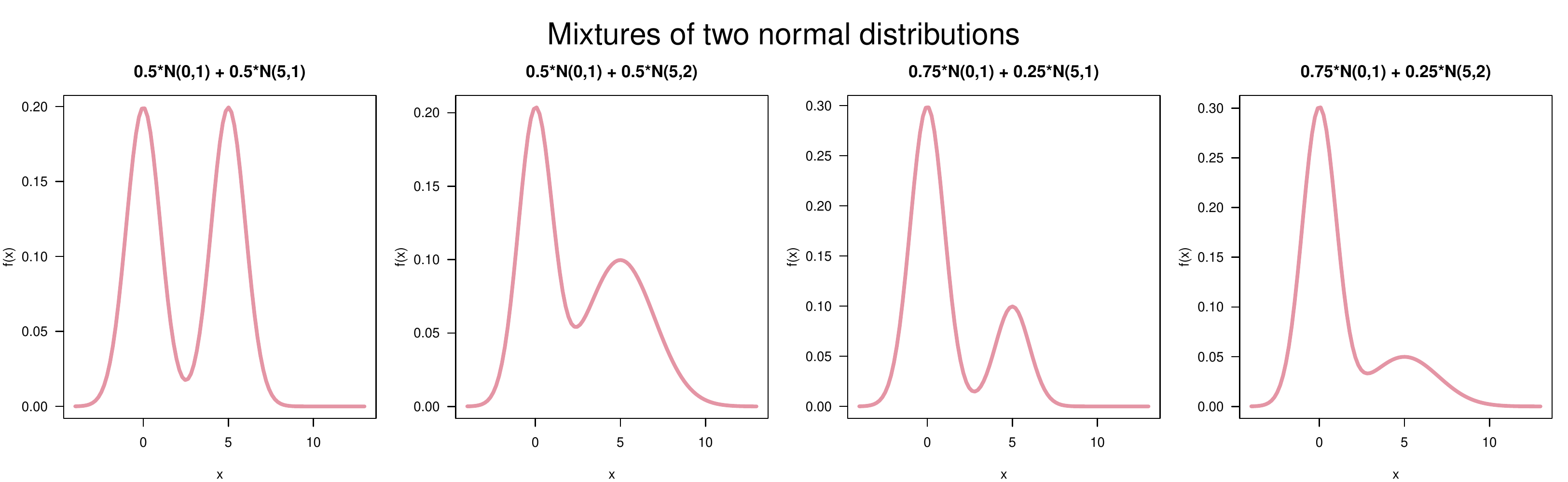}
	\caption{The densities for mixtures of two normal distributions}
	\label{fig:mixnorm_dist_densities}
\end{figure}

Lastly, bimodal distributions are compared in simulation. These are mixture models of two normal distributions with different means and four combinations of variances and proportions; their densities are presented in Figure \ref{fig:mixnorm_dist_densities}. The simulation results given in Figure \ref{fig:normmix_dist_simu_LR2_LR2X} reveal that if $p$ is located in the ``middle point" between the two modes, then the performance of $\hat p_X$ will outperform $\hat p_B$, indicating that staying in the original scale of the $X$-scale will help yield the best results in such cases. However, this performance gain is offset by a large MAE for $\hat p_X$ if $p$ happens to be near a mode.

\subsection{Transforming the predictor variable when its distribution is unknown}

It is often the case, in real-world data, that the distribution of the predictor variable is not known and should (if possible) be estimated from labeled and (if possible) unlabeled observations. Once estimated, the predictor variable(s) could be transformed to $U(0,1)$ using the quantile transformation ($F(X)$) for improving the interpolated misclassification error. Some previous studies have already proposed semi-supervised methods for improving decision trees (such as \cite{criminisi2012decision, tanha2015semi}) but not for improving the split-point interpolation error.

\begin{figure}[h]
	\centering
	\includegraphics[width=0.899\linewidth]{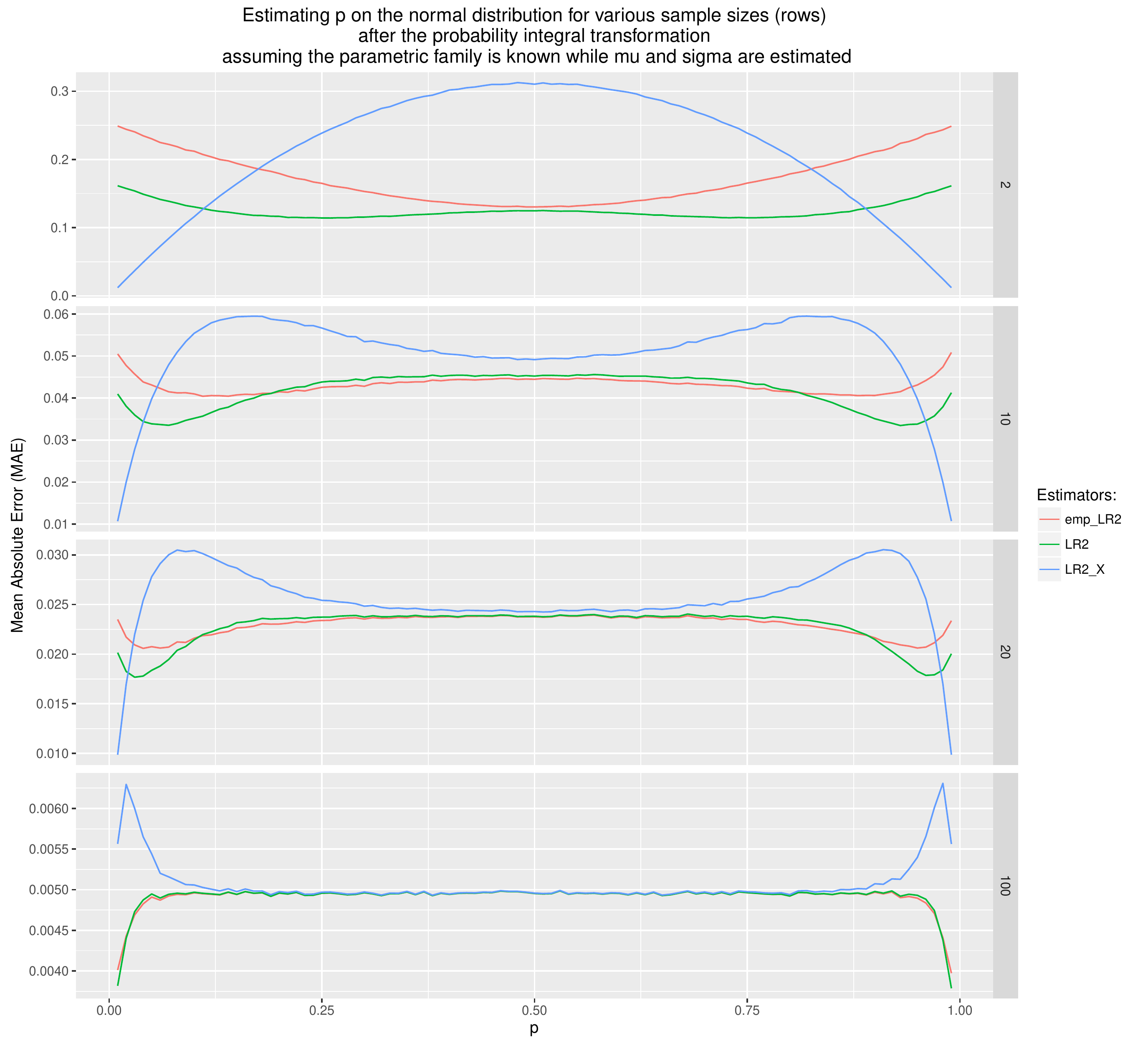}
	\caption{The misclassification error (Mean Absolute Error) of a model estimating $p$ using the original data which comes from a standard normal distribution (blue), estimation after the quantile transformed data (green), and after assuming the distribution is normal and using the estimated $\mu$ and $\sigma$ to estimating the exact distribution. This is done for a range of $p$ positions and sample sizes (2, 10, 20, 100). In the figure $LR2=\hat p_{B}$, $LR2_X=\hat p_X$, and $emp_{LR2}$ refers to using $\hat p_{B}$ using the CDF with the estimated parameters.}
	\label{fig:norm_musigma_est_simu_LR2_LR2X_norm}
\end{figure}

Sometimes the observations come from a postulated parametric family (such as the Normal Distribution), but the exact parameters need to be estimated from the data (e.g. $\mu, \sigma$). In the following simulation the observations are normally distributed, and the quantile transformation applied is based on estimated parameters. Figure \ref{fig:norm_musigma_est_simu_LR2_LR2X_norm} demonstrates how powerful (an adequate!) parametric assumption can be. Already by estimating $\mu$ and $\sigma$ on the labeled data, the estimated quantile transformation (red line) performs almost as well as the true one (green line), both outperforming $X$-scale methods (blue line) in prediction accuracy. 

Even when a normal assumption cannot be justified, for some unimodal data it might be reasonable to assume normality after performing a Box-Cox \citep{box1964analysis} or a Yeo-Johnson  \citep{yeo2000new} transformation (for example, using the caret R package \citep{kuhn2008caret}). The unimodality of the distribution can be tested (for example, using the dip test \citep{hartigan1985dip, pkg_diptest}), and the normality could be tested as well (using shapiro or Kolmogorov-Smirnov tests for normality \citep{shapiro1965analysis, lilliefors1967kolmogorov}). 

When no parametric assumptions are made, $F$ could be estimated using the empirical CDF. Simulation results, presented in Figure \ref{fig:ECDF_semisupervised_simu_LR2_LR2X_norm} and Figure \ref{fig:ECDF_semisupervised_simu_LR2_LR2X_exp} from the appendix, indicate that $n^2$ unlabeled observations from $X$ seem sufficient in order to yield an empirical CDF that is precise enough for building a decision tree model based on $n$ labeled observations. In such cases, the empirical quantile transformation of $X$ could give the superior performance of $\hat p_B$ over $\hat p_X$ (as presented in the following case study). But these figures also show that estimating the CDF by adding only a few unlabeled observations to the labeled ones, is detrimental to performance. Other semi-parametric smoothing kernels may also prove to be useful.

\subsection{Case study - Deciding if a day is rainy or not}

The \textit{weatherAUS} dataset comes bundled with the rattle R package \citep{R_rattle}. The data includes 35,000 daily observations from over 45 Australian weather stations. The data was originally obtained from the Australian Commonwealth Bureau of Meteorology and processed to create a sample dataset for illustrating data mining using R and Rattle.
The data in the package has been processed to provide a target variable \textit{RainTomorrow} on whether there is rain on the following day (No/Yes) and also includes a risk variable \textit{RISK\_MM} which is the amount of rain recorded during the next day. The variable \textit{RISK\_MM} is strictly positive, with a sharply declining, right tailed density function (similar to an Exponential Distribution). As is expected, when the amount of rain in a given day is 0, the \textit{RainTomorrow} variable will indicate that there was no rain. However, a positive amount of measured rain does not necessarily mean that a day would be classified as rainy. In fact, the data shows that a day would be declared rainy only when observations had more than 1.1 (or any location between 1.0 and 1.2) units of rain, and otherwise it would be labeled that there was no rain that day. This split position ($p$) is in the 78\% quantile of the distribution. Could a sample smaller than 35,000 observations be enough for making this distinction? A simulation experiment was conducted based on the dataset; observations were sampled from the full dataset, including the amount of rain (RISK\_MM) and whether the day was rainy or not. The sample was used for finding a split rule, and then the complete dataset of 35,000 observations was used to check the misclassification error of the prediction. This was repeated $10^5$ times, every time with a new sample of observations. The split rule and misclassification errors were averaged over the simulation runs. This was repeated when using the original scale of X (\textit{RISK\_MM}), after using the empirical CDF based on all 35,000 observations (the quantile transformation), and also by using $L$ and $R$. 

\newcommand{\specialcell}[2][c]{%
	\begin{tabular}[#1]{@{}c@{}}#2\end{tabular}}

\begin{table}
	\centering
	
	
	\begin{tabular}{lcccccccc}
		\specialcell{Number of \\ observations} & \multicolumn{4}{c}{\specialcell{Average estimated\\ split point}} & \multicolumn{4}{c}{\specialcell{Interpolated \\ Misclassification error}}\\
		\cmidrule(lr){2-5} 
		\cmidrule(lr){6-9}
		& $X$ & $U$ & $L$ & $R$  & $X$ & $U$ & $L$  & $R$ \\
		\midrule
		10 &  2.65 & 2.62 & 0.40 & 4.89 & 0.0417 & \textbf{0.0409} & 0.0608 & 0.0640 \\ 
		\addlinespace
		20 &  1.70 & 1.67 & 0.60 & 2.80 & 0.0241 & \textbf{0.0233} & 0.0350 & 0.0423 \\ 
		\addlinespace
		100 &  1.12 & 1.06 & 0.93 & 1.31 & 0.0021 & \textbf{0.0019}  & 0.0045 & 0.0056\\ 
		\addlinespace
		1000 &  1.1 & 1.0 & 1.0 & 1.2 & 0.0 & 0.0 & 0.0 & 0.0 \\ 
		\bottomrule
	\end{tabular}
	
		\caption{The effect of the quantile transformation on prediction whether or not it would rain using the amount of measured rain (RISK\_MM). The simulation tested a sample of 10, 20, 100, and 1000 labeled observations from 35,000 daily rainfall observations. The estimation of the CDF used all 35,000 observations (ignoring their labels). $10^5$ samples were drawn for the estimations. In $X$ the estimation is done in the original scale, in $U$ the scale is after the quantile transformation, while $R$ and $L$ uses only the Left and Right locations for estimation.}
		\label{table:weather1}	
		
\end{table}

The results, given in Table \ref{table:weather1}, show that the proposed scenario, in which there is a wealth of unlabeled observations for estimating the CDF, can be leveraged for improving prediction based on only 10, 20 or 100 labeled observations. It is clear that using either $X$ or $U$ is superior to $L$ and $R$, without knowing the exact distribution of $X$. If the CDF of $X$ is known, than there is a small gain to be made when using $U$ instead of $X$. However, once the number of labeled observations is 1000 then all estimators  give perfect results. This might be explained in that while there are 35,000 observations, there are only 477 unique values in the sample. This hints that the observations have been rounded, and using 1000 observations is already enough to represent all the precision offered by data into the ECDF of the underlying distribution.

%
%
%

\section{Discussion}

This study offered several actionable items for both authors and users of statistical algorithms that rely on decision tree models. For writers of such statistical software, the theoretical work in section (\ref{theory_est_p}) leads to the recommendation to use the minimal sufficient statistic $(L,R)$ for split point interpolation - preferably via $\hat p = \frac{{L + R}}{2}$ (or more practically, ${\hat p_{SB}} = \frac{{\hat{p}_{SL} + \hat{p}_{SR}}}{2}$). Even when the distribution of $X$ is not uniform, this split point interpolation will yield better results than only using $L_X$ or $R_X$ for predicting observations with new values. This (average) improvement of interpolated predictions will sometimes take place for discrete variables but will always be relevant for continuous predictor variables.

For users of decision tree algorithms, the simulation results in section (\ref{beyond_SU_seq}) indicate that knowledge about the distribution of $X$ can be used to improve the prediction of the algorithm on future observations (if the DTL uses $\hat p_X = \frac{{L_X + R_X}}{2}$), by using the quantile transformation $X$ to get $\hat p_{B} = {{F(L_X)+F(R_X)} \over 2} = {{L+R} \over 2} $. The most significant gain for transforming $L_X$ and $R_X$ to $L$ and $R$ occurs when the number of observations used for deciding on a split is small. Transforming $X$ to the quantile scale will be most beneficial when the density near the split is skewed (thus, making the median of the posterior distribution close to one of the edges of either $L_X$ or $R_X$). The method is less likely to be helpful if $X$ is essentially separated by $p$ into two clusters (see the mixture cases in Figure \ref{fig:normmix_dist_simu_LR2_LR2X}).

When the distribution of $X$ is not known, adequate parametric modeling may permit improvement in prediction accuracy even when estimation is based on the labeled data exclusively (see Figure \ref{fig:norm_musigma_est_simu_LR2_LR2X_norm}, even for $n=10$). 

While this study focused on a single split for a binary deterministic response variable ($Y$), the results are in fact indicative to any type of recursive binary decision tree, be it a multi-class problem or a regression problem. A recursive binary decision tree is such that at every node the split point interpolation problem is the same as we have dealt with, in the sense that one $p$ is estimated at a time (for this subset of the data). The limitation is that now the support of the observations is not fixed (but may depend on previous splits), but if the split is away from the support of the conditional distribution (and given that there are observations from both sides of $p$) then it should not influence the conclusions in this study. As for non-binary response, for example, if $X \sim U(0,1)$ and $Y|X<p \sim N(0,1)$ while $Y|X>p \sim N(10,1)$ then the split point interpolation problem of estimating $p$ is still similar to everything that was discussed until now. The main difference is that the performance might use a different metric than the misclassification error used in this paper. When the decision tree model uses multiple splits, as is often the case, the benefit of using the quantile transformation on the predictor variables depends on the relation between the number of observations and the complexity of the estimated model. See section \ref{sec:more_than_one_split} in the appendix for some simulation results. 

Lastly, it is often the case in real-world problems that the $0-1$ variable $Y$ (rather than deterministic) is a stochastic monotone function of $X$, in which case the pair $(L,R)$ is not well defined, and the supervised uniform distribution model should undergo some isotonic generalization. This is a subject for further research.


\acks{
We thank Professor Yoav Benjamini, Professor Eilon Solan, Professor Dafna Shahaf, Professor Saharon Rosset,  Yoni Sidi, Barak Brill, Marilyn Friedes, and Deborah Galili for their valuable input on this manuscript. This work was supported in part by the European Research Council under EC–EP7 European Research Council grant PSARPS-297519, and in part by the European Union Seventh Framework Programme (FP7/2007-2013) under grant agreement no. 604102 (Human Brain Project).}




\vskip 0.2in
\bibliography{DT_bib}



\newpage

\section{Appendix}

\subsection{Introduction to Decision Tree Models and CART}
\label{app:DTL_intro}

\subsubsection{Introduction to Decision Tree Models}

Decision tree learning (DTL) is any algorithm which constructs a decision tree model (DTM) as a predictive model for mapping observations about an item (also known as explanatory or predictor variables) to conclusions about the item's target value (also known as dependent or response variable). Decision tree model (or classifier) are very commonly used for predictive modeling in statistics, data mining and machine learning. DTL is often performed by searching for a series of decision rules which partition the space spanned by the independent variables into disjoint region of the space. Each partitioned region of the space is attributed with some prediction of the dependent variable. In these tree structures, leaves represent the predicted target and branches represent conjunctions of features that lead to those predictions. Decision trees where the target variable can take continuous values (typically real numbers) are called regression trees, and when the target variable can take a finite set of values the models are called classification trees. Generally speaking, decision trees are primarily intended as prediction models for cases where we have no preconceived notion about the structure of the model that would fit the data. The problem of learning an optimal decision tree is known to be NP-complete under several aspects of optimality \citep{hyafil1976constructing}. Hence, many DTL methods use a mixture of randomization and greedy (e.g. forward step-wise) approaches for searching the data space, in the hopes of stumbling upon a "useful" patterns for the prediction of interest. In order to protect from over-fitting to the training data at hand, DTL algorithms employ various methods for restricting the model's complexity while optimizing some measure of prediction accuracy.

Research on DTL dates back to works in the social sciences from the 60's by Morgan and Sonquist \citep{morgan1963problems} which was later improved in the 80's by Breiman, et al. “Classification and regression trees” (CART) methodology \citep{breiman1984classification}, and in the 90's by Quinlan's C4.5 \citep{quinlan1993c4}. As time passed, DTL started to be used in more fields, and today they are applied in multiple disciplines such as statistics, pattern recognition, decision theory, signal processing, machine learning and artificial neural networks. A survey of decision trees’ adaptation is provided by S. K. Murthy \citep{murthy1998automatic}, and a more recent survey by S Lomax and S Vadera concluded that, in 2013, there has already been over 50 different algorithms for producing decision trees (including ID3, CHAID, C5.0, oblique trees, etc.) \citep{lomax2013survey}. This work will use the CART DTL methodology for illustration and for some of the simulations, although the conclusion from this work are equally applicable to most of the other algorithms.

\subsubsection{Introduction to CART}

Depicted in 2008 as one of the top ten algorithms in data mining \citep{wu2008top}, CART \citep{breiman1984classification} offers a good example for a DTL. This section outlines the major steps in the CART algorithm. The construction of the decision tree starts in the root node which consists of the entire learning set, then (1) all possible variables are scanned for possible splits, and (2) the variable+split with the best "impurity" measure is picked - possible impurity measures can be misclassification, Gini or entropy for classification trees and mean square error for regression trees. The root node is split into two nodes by choosing one of the explanatory variables and making a rule on the chosen predictor variable which divides it into two groups (based on steps 1 and 2). Then (3) steps 1 and 2 are recursively repeated for each of the child nodes until a predefined stopping rule is met for all terminal nodes (e.g., pre-pruning, or a node reached a certain minimal number of observations), (4) each node is assigned a prediction by funneling the training dataset to it, and picking the majority class for nominal dependent variables or the average for numerical variables. From the complete tree (5) a series of nested (pruned) sub-trees are defined based on misclassification error when using the training data-set, (6) a k-fold cross validation (CV) is performed by which the above process is repeated k times (each time on a fraction of $\frac{k-1}{k}$ the sample size), (7) using the CV hold-out samples on their respective models (each on a fraction $\frac{1}{k}$ of the sample size), a complexity parameter is determined so that the cross-validated error is minimized, and the entire tree is pruned by cost-complexity trade-off. The prediction of a new observation is made by funneling it through the nodes (based on their corresponding values in the explanatory variables) until it falls into one of the terminal nodes (i.e., a leaf node which has no splits), where it is given a prediction based on some aggregate of values from the training set (see step (4) above).

While the CART methodology offers a useful search mechanism in the model space, it has several known limitations when used for prediction. Small CART trees tend to give biased predictions while large trees have high variance and tend to over-fit the data. While CART's cost-complexity pruning offers a reasonable compromise on tree size, better alternatives have since been presented by extending a single decision tree to an ensembles of decision trees. Leo Breiman is responsible for two celebrated extensions of CART - Bagging \citep{breiman1996bagging} and Random Forest \citep{breiman2001random}. Both of these methods draw many bootstrap samples from the training data in order to grow full-sized (un-pruned) decision trees – and then aggregate their predictions (majority vote for classification or averaging for regression trees). Bagging samples can choose among all the potential features when deciding on a split while random forest chooses at each split only among a randomly chosen subset of the features. Another (intermediate) method is the random subspace method by Ho \citep{ho1998random}, which samples the features only once for each of the bagged samples (instead of at each node, as is done in random forest). These extensions search a wider range of models, and by combining large trees with aggregation over many trees they reduce the bias and variance of the model - often yielding superior predictive performance over CART. Another prominent alternative, suggested by Freund and Schapire \citep{freund1995desicion}, is the idea of boosting as implemented in algorithms such as AdaBoost. Comparison of these methods has shown that each may be superior to others in alternative scenarios \citep{banfield2007comparison}. Further extensions include gradient boosting and others.

\newpage

\subsection{Integrals for section \ref{sec:SU_p_estimation}}
\label{app:integrals}

\subsubsection{${{\hat p}_{RB}}$}
\label{app:p_RB_extra_equations}

As an exercise, ${{\hat p}_{RB}}$ is indeed unbiased:
\begin{align}
E_p\left[ {{{\hat p}_{RB}}} \right] &= E\left[ {\left[ {\frac{1}{n} + \frac{{n - 2}}{n}\frac{L}{{L + 1 - R}}} \right]{I_{\left\{ {0 < L<R < 1} \right\}}} + {I_{\left\{ {R = 1} \right\}}}} \right]\nonumber \\
&= {\frac{1}{n}E\left( {{I_{\left\{ {0 < L<R < 1} \right\}}}} \right) + \frac{{n - 2}}{n}E\left[{\frac{L}{{L + 1 - R}}{I_{\left\{ {0 < L<R < 1} \right\}}}} \right]} + E\left[ {{I_{\left\{ {R = 1} \right\}}}} \right]\nonumber \\
&= \frac{1}{n}P\left( {0 < L<R < 1} \right)+ P\left( {R = 1} \right) \nonumber \\
&+ (n-1)(n-2)\int_0^p {\int_p^1 {\left( {\frac{l}{{l + 1 - r}}} \right){{\left( {l + 1 - r} \right)}^{n - 2}}dr} dl} 
\nonumber \\
&= \frac{1}{n}\left[ {1 - {p^n} - {{\left( {1 - p} \right)}^n}} \right] + {p^n} + \frac{{{p^n} + n\left( {p - {p^n}} \right) + {{(1 - p)}^n} - 1}}{{n}}= p \label{EXPRB} 
\end{align}

Needed calculations for the variance of ${{\hat p}_{RB}}$

\begin{align}
E\left[ {{{\hat p}_{RB}}^2} \right] &= E\left[ {{{\left[ {\frac{1}{n} + \frac{{n - 2}}{n}\frac{L}{{L + 1 - R}}} \right]}^2}{I_{\left\{ {0 < L<R < 1} \right\}}} + {I_{\left\{ {R = 1} \right\}}}} \right] \nonumber\\
&= E\left[ {\frac{1}{{{n^2}}}\left[ {1 + 2\left( {n - 2} \right)\frac{L}{{L + 1 - R}} + {{\left( {n - 2} \right)}^2}{{\left( {\frac{L}{{L + 1 - R}}} \right)}^2}} \right]{I_{\left\{ {0 < L<R < 1} \right\}}} + {I_{\left\{ {R = 1} \right\}}}} \right] \nonumber\\
&= \frac{1}{{{n^2}}}\left[ {\begin{array}{*{20}{l}}
	{E\left( {{I_{\left\{ {0 < L<R < 1} \right\}}}} \right) + 2\left( {n - 2} \right)E\left( {\frac{L}{{L + 1 - R}}{I_{\left\{ {0 < L<R < 1} \right\}}}} \right)}\\
	{ + {{\left( {n - 2} \right)}^2}E\left( {{{\left( {\frac{L}{{L + 1 - R}}} \right)}^2}{I_{\left\{ {0 < L<R < 1} \right\}}}} \right)}
	\end{array}} \right] + E\left[ {{I_{\left\{ {R = 1} \right\}}}} \right] \nonumber\\
&= \frac{1}{{{n^2}}}\left[ {\begin{array}{*{20}{l}}
	{P\left( {0 < L<R < 1} \right) + 2\left( {n - 2} \right){I_{\left\{ {0 < L<R < 1} \right\}}}\int_0^p {\int_p^1 {\left( {\frac{l}{{l + 1 - r}}} \right){f_{L,R}}\left( {l,r} \right)dl} dr} }\\
	{ + {{\left( {n - 2} \right)}^2}{I_{\left\{ {0 < L<R < 1} \right\}}}\int_0^p {\int_p^1 {{{\left( {\frac{l}{{l + 1 - r}}} \right)}^2}{f_{L,R}}\left( {l,r} \right)dl} dr} }
	\end{array}} \right] + P\left( {R = 1} \right) \nonumber\\
&= \frac{1}{{{n^2}}}\left[ {\begin{array}{*{20}{l}}
	{\left[ {1 - {p^n} - {{\left( {1 - p} \right)}^n}} \right] + 2\left( {n - 2} \right)\int_0^p {\int_p^1 {\left( {\frac{l}{{l + 1 - r}}} \right)n\left( {n - 1} \right){{\left( {l + 1 - r} \right)}^{n - 2}}dr} dl} }\\
	{ + {{\left( {n - 2} \right)}^2}\int_0^p {\int_p^1 {{{\left( {\frac{l}{{l + 1 - r}}} \right)}^2}n\left( {n - 1} \right){{\left( {l + 1 - r} \right)}^{n - 2}}dr} dl} }
	\end{array}} \right] + {p^n}
\end{align}

For calculating $E\left[ {{{\hat p}_{RB}}^2} \right]$ the following needs to be calculated $ \int_0^p {\int_p^1 {{{\left( {\frac{l}{{l + 1 - r}}} \right)}^2}{{\left( {l + 1 - r} \right)}^{n - 2}}dr} dl} $. This should be separated into two cases:






\begin{align}
n = 3 &\Rightarrow \int_0^p {\int_p^1 {{{\left( {\frac{l}{{l + 1 - r}}} \right)}^2}{{\left( {l + 1 - r} \right)}^{n - 2}}dr} dl} \nonumber \\
&= \frac{1}{6}\left( { - 2\left[ {{{(1 - p)}^3}\log (1 - p) + {p^3}\log (p)} \right] - p\left( {1 - p} \right)\left( {2\left( {1 - p} \right) - p} \right)} \right) \\ 
n \ne 3 &\Rightarrow \int_0^p {\int_p^1 {{{\left( {\frac{l}{{l + 1 - r}}} \right)}^2}{{\left( {l + 1 - r} \right)}^{n - 2}}dr} dl}  = \frac{{\frac{{ - 2{{(1 - p)}^n} + np((n - 1)p - 2) + 2}}{{(n - 2)(n - 1)}} - {p^n}}}{{(n - 3)n}}
\end{align}


For $n \ne 3$, this is the expectation:


\begin{align}
E\left[ {{{\hat p}_{RB}}^2} \right] &= \nonumber \\
&= \frac{1}{{{n^2}}}\left[ \begin{array}{l}
\left[ {1 - {p^n} - {{\left( {1 - p} \right)}^n}} \right] + 2\left( {n - 2} \right)n\left( {n - 1} \right)\frac{{{p^n} + n\left( {p - {p^n}} \right) + {{(1 - p)}^n} - 1}}{{(n - 2)(n - 1)n}} \nonumber\\
+ {\left( {n - 2} \right)^2}n\left( {n - 1} \right)\frac{{\frac{{ - 2{{(1 - p)}^n} + np((n - 1)p - 2) + 2}}{{(n - 2)(n - 1)}} - {p^n}}}{{(n - 3)n}}
\end{array} \right] + {p^n} \nonumber\\
&= \frac{{\left( {1 - n} \right)\left[ {{p^n} + {{(1 - p)}^n} - 1} \right] - 2np\left( {1 - p} \right)}}{{(n - 3){n^2}}} + {p^2}
\end{align}

And $n = 3$ gets

\begin{align}
E\left[ {{{\hat p}_{RB}}^2} \right] &= \nonumber\\
&= \frac{1}{9}\left[ \begin{array}{l}
\left[ {1 - {p^3} - {{\left( {1 - p} \right)}^3}} \right] + 2\left( {{p^3} + 3\left( {p - {p^3}} \right) + {{(1 - p)}^3} - 1} \right)\nonumber\\
+ \left( { - 2\left[ {{{(1 - p)}^3}\log (1 - p) + {p^3}\log (p)} \right] - p\left( {1 - p} \right)\left( {2\left( {1 - p} \right) - p} \right)} \right)
\end{array} \right] + {p^3}\nonumber\\
&= \frac{1}{9}\left( {p\left( {1 - p} \right) - 2\left( {{p^3}\log (p) + {{(1 - p)}^3}\log (1 - p)} \right)} \right) + {p^2}
\end{align}

For $n \ne 3$, the variance of ${\hat p}_{RB}$ is 
\begin{align} \label{RB_variance_extended}
V_p^{(n \ne 3)}\left[ {{{\hat p}_{RB}}} \right] &= E_p\left[ {{{\hat p}_{RB}}^2} \right] - {E_p^2}\left[ {{{\hat p}_{RB}}} \right] = E\left[ {{{\hat p}_{RB}}^2} \right] - {p^2}\nonumber \\
&= \frac{1}{{(n - 3)n}}\left[ {\frac{{\left( {n - 1} \right)}}{n}\left[ {1 - \left( {{p^n} + {{(1 - p)}^n}} \right)} \right] - 2p\left( {1 - p} \right)} \right] \nonumber \\
&\approx {{1-2p(1-p)} \over n^2}
\end{align}
Specifically, for $n=2$ the variance is
\begin{align}
V_p^{(n=2)}\left[ {{{\hat p}_{RB}}} \right] =  - \frac{1}{2}\left[ { - \frac{1}{2}\left[ {1 - \left( {{p^2} + {{(1 - p)}^2}} \right)} \right] - 2p\left( {1 - p} \right)} \right] = \frac{1}{2}(1 - p)p
\end{align}

For $n=3$ variance evaluation requires a special treatment:

\begin{align}
	V_p^{(n=3)}\left[ {{{\hat p}_{RB}}} \right] 
	&= \frac{1}{9}\left( {p\left( {1 - p} \right) - 2\left( {{p^3}\log (p) + {{(1 - p)}^3}\log (1 - p)} \right)} \right)
	\label{RB_variance_3}
\end{align}

\subsubsection{Estimating $p$ using $L$} \label{sec:estimating_p_using_L}

The expectation of $L$ can be evaluated applying eq. (\ref{E_g_LR}) to $g\left( {L,R} \right) = L$:

\begin{align}
E_p\left[ L \right] &= 0 + n\left( {n - 1} \right)\int_0^p {\int_p^1 {l{{\left( {l + 1 - r} \right)}^{n - 2}}dr} dl}  + n\int_0^p {{l^n}dl}   \nonumber \\
&= n\left( {n - 1} \right)\frac{{\left[ {{{(1 - p)}^{n + 1}} + np + p - 1} \right] - n{p^{n + 1}}}}{{n\left( {n - 1} \right)\left( {n + 1} \right)}} + n\frac{{{p^{n + 1}}}}{{n + 1}} \nonumber \\
&= p - \frac{{1 - {{\left( {1 - p} \right)}^{n + 1}}}}{{n + 1}} \label{EL}
\end{align}

Finding $E(L^2)$

\begin{align} \label{E_L2}
E\left[ {{L^2}} \right] &= 0 + n\left( {n - 1} \right)\int_0^p {\int_p^1 {{l^2}{{\left( {l + 1 - r} \right)}^{n - 2}}dr} dl}  + n\int_0^p {{l^{n + 1}}dl} \nonumber \\
&= \frac{{p\left( {{n^2}( - p)\left( {{p^n} - 1} \right) - n\left( {p\left( {{p^n} - 3} \right) + 2} \right) - 2(p - 2)\left( {{{(1 - p)}^n} - 1} \right)} \right) - 2{{(1 - p)}^n} + 2}}{{(n + 1)(n + 2)}} + n\frac{{{p^{n + 2}}}}{{n + 2}} \nonumber \\
&= \frac{{ - 2(p - 2)p{{(1 - p)}^n} - 2{{(1 - p)}^n} + (n + 2)p(np + p - 2) + 2}}{{(n + 1)(n + 2)}}
\end{align}

%
%
%
%
%
%
%
%
%

And the variance $V_p(L)$ is
\begin{align}
V_p\left[ L \right] &= E_p\left[ {{L^2}} \right] - {E_p^2}\left[ L \right] \nonumber \\
&= \frac{{{p^2}(n + 1)(n + 2) - 2\left( {n + 2} \right)p + 2 - 2{{(1 - p)}^{n + 2}}}}{{(n + 1)(n + 2)}} - {{\left( {p - \frac{{1 - {{\left( {1 - p} \right)}^{n + 1}}}}{{\left( {n + 1} \right)}}} \right)}^2} \nonumber \\
&=	{ \frac{{\begin{array}{*{20}{l}}
			{\left( {2{n^2}(p - 1)p - 2{{(p - 1)}^2}\left( {{{(1 - p)}^n} - 1} \right)} \right){{(1 - p)}^n}} \nonumber \\
			{ + n\left( { - \left( {{p^2} - 2p + 1} \right){{(1 - p)}^{2n}} + 4(p - 1)p{{(1 - p)}^n} + 1} \right)}
			\end{array}}}{{{{(n + 1)}^2}(n + 2)}}} \nonumber \\
&=	{ \frac{{ - 2{{(1 - p)}^{1 + n}}( - ( - 1 + {{(1 - p)}^n})( - 1 + p) + {n^2}p) + n(1 - {{(1 - p)}^{2(1 + n)}} - 4{{(1 - p)}^{1 + n}}p)}}{{{{(n + 1)}^2}(n + 2)}}}
\nonumber \\
&= \frac{{\frac{n}{{\left( {n + 2} \right)}} - 2\left( {np - \frac{1}{{\left( {n + 2} \right)}}\left( {1 - p} \right)} \right){{(1 - p)}^{n + 1}} - {{(1 - p)}^{2(n + 1)}}}}{{{{(n + 1)}^2}}}
\label{VL}
\end{align}
leading to
\begin{align}
MSE_p\left[ L \right] &= 
\frac{{\frac{n}{{\left( {n + 2} \right)}} - 2\left( {np - \frac{1}{{\left( {n + 2} \right)}}\left( {1 - p} \right)} \right){{(1 - p)}^{n + 1}} - {{(1 - p)}^{2(n + 1)}}}}{{{{(n+1)}^2}}}
+ {\left( {\frac{{1 - {{\left( {1 - p} \right)}^{n + 1}}}}{{n + 1}}} \right)^2} \nonumber \\
&= \frac{{2\left( {1 - \left( {p\left( {n + 1} \right) + 1} \right){{(1 - p)}^{n + 1}}} \right)}}{{(n + 1)(n + 2)}} \approx {2 \over {n^2}} \label{MSEL}
\end{align}

\subsubsection{Estimating $p$ using $R$}

 $E_p(R)$ and $V_p(R)$ are given by
 \begin{align}
 E_p\left[ R \right] &= n\int_p^1 {r{{\left( {1 - r} \right)}^{n - 1}}dr}  + n\left( {n - 1} \right)\int_0^p {\int_p^1 {r{{\left( {l + 1 - r} \right)}^{n - 2}}dr} dl}  + n\int_0^p {{l^{n - 1}}dl} \nonumber \\
 &= p + \frac{{1 - {p^{n + 1}}}}{{ {n + 1} }} \label{ER}
 \end{align}
 and
 \begin{align}
 V_p\left[ R \right] &= E_p\left[ {{R^2}} \right] - {E_p^2}\left[ R \right] \nonumber \\
 &= \frac{{\frac{n}{{\left( {n + 2} \right)}} - 2\left( {n(1 - p) - \frac{1}{{\left( {n + 2} \right)}}p} \right){p^{n + 1}} - {p^{2(n + 1)}}}}{{{{(n + 1)}^2}}}
 \label{VR}
 \end{align}
 leading to
\begin{align}
 MSE_p\left[ R \right] &=
 \frac{{\frac{n}{{\left( {n + 2} \right)}} - 2\left( {n(1 - p) - \frac{1}{{\left( {n + 2} \right)}}p} \right){p^{n + 1}} - {p^{2(n + 1)}}}}{{{(n + 1)^2}}}
 + {\left( {\frac{{1 - {p^{n + 1}}}}{n + 1}} \right)^2} \nonumber \\
 &= \frac{{2\left( {1 - ((1 - p)\left( {n + 1} \right) + 1){p^{n + 1}}} \right)}}{{(n + 1)(n + 2)}} \approx {2 \over {n^2}} \label{MSER}
 \end{align}

Needed calculations:

1) 


\begin{align}
\int_p^1 {r{{\left( {1 - r} \right)}^{n - 1}}dr}  = \frac{{{{\left( {1 - p} \right)}^n}\left( {np + 1} \right)}}{{n\left( {n + 1} \right)}}
\end{align}

2)


\begin{align}
\int_0^p {\int_p^1 {r{{\left( {l + 1 - r} \right)}^{n - 2}}dr} dl}  = \frac{{ - {p^n}(n + p + 1) - (np + 1){{(1 - p)}^n} + np + p + 1}}{{n\left( {{n^2} - 1} \right)}}
\end{align}


Finding $E(R^2)$


\begin{align}
E\left[ {{R^2}} \right] &= n\int_p^1 {{r^2}{{\left( {1 - r} \right)}^{n - 1}}dr}  + n\left( {n - 1} \right)\int_0^p {\int_p^1 {{r^2}{{\left( {l + 1 - r} \right)}^{n - 2}}dr} dl}  + n\int_0^p {{l^{n - 1}}dl} \nonumber\\
&= \frac{{{{(1 - p)}^n}(np(np + p + 2) + 2)}}{{(n + 1)(n + 2)}} + \nonumber\\
&+ \frac{{ - {p^n}\left( {{n^2} + n(2p + 3) + 2{{(p + 1)}^2}} \right) - (np(np + p + 2) + 2){{(1 - p)}^n} + (n + 2)p(np + p + 2) + 2}}{{(n + 1)(n + 2)}}\nonumber\\
&+ {p^n}\nonumber\\
&= {p^2} - \frac{{2\left( {2{p^{n + 1}} + {p^{n + 2}} + n\left( {{p^{n + 1}} - p} \right) - 2p - 1} \right)}}{{(n + 1)(n + 2)}}
\end{align}

%
%
%
%
%
%
%
%
%
%

%
%
%

%
%
%
%



\subsubsection{Estimating $p$ using $(L+R)/2$}


Finding the variance requires several calculations

\begin{align}
{E^2}\left[ {\frac{{L + R}}{2}} \right] &= {\left( {p + \frac{{{{\left( {1 - p} \right)}^{n + 1}} - {p^{n + 1}}}}{{2\left( {n + 1} \right)}}} \right)^2} \nonumber \\
&= \frac{{4{{\left( {n + 1} \right)}^2}{p^2} + 4\left( {n + 1} \right)p\left( {{{\left( {1 - p} \right)}^{n + 1}} - {p^{n + 1}}} \right)+{{\left( {{{\left( {1 - p} \right)}^{n + 1}} - {p^{n + 1}}} \right)}^2}}}{{4{{\left( {n + 1} \right)}^2}}}
\end{align}

Also
\begin{align}
&E\left[ {{{\left( {L + R} \right)}^2}} \right] \nonumber\\
&= n\int_p^1 {{r^2}{{\left( {1 - r} \right)}^{n - 1}}dr}  + n\left( {n - 1} \right)\int_0^p {\int_p^1 {{{\left( {l + r} \right)}^2}{{\left( {l + 1 - r} \right)}^{n - 2}}dr} dl}  + n\int_0^p {{{\left( {u + 1} \right)}^2}{l^{n - 1}}dl} \nonumber\\
&= n\frac{{{{(1 - p)}^n}(np(np + p + 2) + 2)}}{{n(n + 1)(n + 2)}}\nonumber\\
&+ n\left( {n - 1} \right)\frac{{\left( {n + 2} \right)\left( {n + 1} \right)\left( {\left( {4 - \left( {{{\left( {1 - p} \right)}^n} + {p^n}} \right)} \right){p^2} - {p^n}\left( {1 + 2p} \right)} \right) + 2 - 2\left( {{p^{n + 2}} + {{\left( {1 - p} \right)}^{n + 2}}} \right)}}{{(n - 1)n(n + 1)(n + 2)}}\nonumber\\
&+ n\frac{{\left( {n + 1} \right){p^{n + 2}}n + \left( {n + 2} \right)2{p^{n + 1}}n + \left( {n + 1} \right)\left( {n + 2} \right){p^n}}}{{\left( {n + 1} \right)\left( {n + 2} \right)n}}
\end{align}

This simplifies to:

\begin{align}
E\left[ {{{\left( {L + R} \right)}^2}} \right] = \frac{{2\left( {1 - (n + 2)\left( {p\left( {{p^n} - {{(1 - p)}^n}} \right) + {p^2}\left( {{p^n} + {{(1 - p)}^n} - 2\left( {n + 1} \right)} \right)} \right)} \right)}}{{(n + 1)(n + 2)}}
\end{align}

The variance is
\begin{align}
V_p\left[ {\frac{{L + R}}{2}} \right] &= \left( {\frac{1}{4}E\left[ {{{\left( {L + R} \right)}^2}} \right] - {E^2}\left[ {\frac{{L + R}}{2}} \right]} \right)  \nonumber \\
&= \frac{{\left( {1 - (n + 2)\left( {p\left( {{p^n} - {{(1 - p)}^n}} \right) + {p^2}\left( {{p^n} + {{(1 - p)}^n} - 2\left( {n + 1} \right)} \right)} \right)} \right)}}{{2(n + 1)(n + 2)}}\nonumber \\
&- \frac{{4{{\left( {n + 1} \right)}^2}{p^2} + 4\left( {n + 1} \right)p\left( {{{\left( {1 - p} \right)}^{n + 1}} - {p^{n + 1}}} \right) + {{\left( {{{\left( {1 - p} \right)}^{n + 1}} - {p^{n + 1}}} \right)}^2}}}{{4{{\left( {n + 1} \right)}^2}}} \nonumber \\
&= \frac{{1 - (n + 2)p\left( {1 - p} \right)\left( {{p^n} + {{(1 - p)}^n}} \right)}}{{2(n + 1)(n + 2)}} - \frac{{{{\left( {{{(1 - p)}^{n + 1}} - {p^{n + 1}}} \right)}^2}}}{{4{{(n + 1)}^2}}} \nonumber \\ 
&\approx {1 \over {2n^2}} 
\label{VarBayes}
\end{align}
from which its MSE is
\begin{align}
MSE_p\left[ {\frac{{L + R}}{2}} \right] &= \frac{{1 - (n + 2)p\left( {1 - p} \right)\left( {{p^n} + {{(1 - p)}^n}} \right)}}{{2(n + 1)(n + 2)}} \label{MSEBayes} \\					
&- \frac{{{{\left( {{{(1 - p)}^{n + 1}} - {p^{n + 1}}} \right)}^2}}}{{4{{(n + 1)}^2}}} 
+ {\left( {\frac{{{{\left( {1 - p} \right)}^{n + 1}} - {p^{n + 1}}}}{{2\left( {n + 1} \right)}}} \right)^2} \nonumber \\
&= \frac{{1 - (n + 2)(1 - p)p\left( {{p^n} + {{(1 - p)}^n}} \right)}}{{2(n + 1)(n + 2)}} \approx {1 \over {2n^2}} \nonumber
\end{align}
Variance and MSE are symmetric around $p=0.5$, as expected.

This Mean Absolute Error function is minimized by taking the median of the posterior (uniform) distribution, which, yet again, is ${{\hat p}_{B}} = \frac{{L + R}}{2}$.


\begin{align}
MA{E_p}\left[ {\frac{{L + R}}{2}} \right] &= E\left[ {\left| {\frac{{L + R}}{2} - p} \right|} \right] \nonumber\\
&= n\int_p^1 {\left| {\frac{1}{2}r - p} \right|{{\left( {1 - r} \right)}^{n - 1}}dr}  \nonumber \\
&+ n\left( {n - 1} \right)\int_0^p {\int_p^1 {\left| {\frac{{l + r}}{2} - p} \right|{{\left( {l + 1 - r} \right)}^{n - 2}}dr} dl}  \nonumber \\
&+ n\int_0^p {\left| {\frac{{l + 1}}{2} - p} \right|{l^{n - 1}}dl} \nonumber \\
&= \frac{{1 - \left( {{p^{n + 1}} + {{\left( {1 - p} \right)}^{n + 1}}} \right) + {{\left( {\left| {p - \left( {1 - p} \right)} \right|} \right)}^{n + 1}}}}{{2(n + 1)}} \approx \frac{1}{2n}
\end{align}

\subsubsection{Sweeping estimators}

The mean of $L_S$:

\begin{align} \label{E_L_S}
E\left[ {{L_S}} \right] &= E\left[ {{I_{\left\{ {R < 1} \right\}}}L + {I_{\left\{ {R = 1} \right\}}}} \right] = \nonumber \\
&= \left( {p - \frac{{1 - {{\left( {1 - p} \right)}^{n + 1}}}}{{n + 1}}} \right)\left( {1 - {p^n}} \right) + {p^n} \nonumber \\
&= p - \frac{{1 - {{\left( {1 - p} \right)}^{n + 1}}}}{{n + 1}} + \left( {1 - p + \frac{{1 - {{\left( {1 - p} \right)}^{n + 1}}}}{{n + 1}}} \right){p^n}
\end{align}

And also for $R_S$:

\begin{align}
{E_p}\left[ {{R_S}} \right] &= 0 + n\left( {n - 1} \right)\int_0^p {\int_p^1 {r{{\left( {l + 1 - r} \right)}^{n - 2}}dr} dl}  + n\int_0^p {{l^{n - 1}}dl} \nonumber \\
&= 0 + \frac{{ - {p^n}(n + p + 1) - (np + 1){{(1 - p)}^n} + np + p + 1}}{{\left( {n + 1} \right)}} + {p^n} \nonumber\\
&= p + \frac{{1 - {p^{n + 1}}}}{{\left( {n + 1} \right)}} - \frac{{\left( {np + 1} \right){{\left( {1 - p} \right)}^n}}}{{\left( {n + 1} \right)}}
\end{align}

Allowing for:

\begin{align}
E\left[ {{{\hat p}_{SB}}} \right] &= E\left[ {\frac{{{L_S} + {R_S}}}{2}} \right] \nonumber\\
&= p + \frac{{{{\left( {1 - p} \right)}^{n + 1}} - {p^{n + 1}}}}{{2\left( {n + 1} \right)}} + \frac{1}{2}\left( {{p^n}\left( {1 - \frac{n}{{n + 1}}p} \right) - \frac{{\left( {np + 1} \right){{\left( {1 - p} \right)}^n}}}{{\left( {n + 1} \right)}}} \right) \nonumber\\
&= p + \frac{1}{2}\left( {{p^n}\left( {1 - p} \right) - p{{\left( {1 - p} \right)}^n}} \right)
\end{align}

The needed calculations for the MSE:

\begin{align}
{E\left[ {{{\left( {{L_S} + {R_S}} \right)}^2}} \right]} &={ 0 + n\left( {n - 1} \right)\int_0^p {\int_p^1 {{{\left( {l + r} \right)}^2}{{\left( {l + 1 - r} \right)}^{n - 2}}dr} dl}  + 4{p^n}} \nonumber \\
&={ \left( {\left( {4 - \left( {{{\left( {1 - p} \right)}^n} + {p^n}} \right)} \right){p^2} - {p^n}\left( {1 + 2p} \right)} \right) + \frac{{2\left( {1 - \left( {{p^{n + 2}} + {{\left( {1 - p} \right)}^{n + 2}}} \right)} \right)}}{{(n + 1)(n + 2)}} + 4{p^n}} \nonumber \\
&={ 4{p^2} - {{\left( {1 - p} \right)}^n}{p^2} + {p^n}\left( {1 - p} \right)\left( {p + 3} \right) + \frac{{2\left( {1 - \left( {{p^{n + 2}} + {{\left( {1 - p} \right)}^{n + 2}}} \right)} \right)}}{{(n + 1)(n + 2)}}}
\end{align}

The variance is:

\begin{align}
&V\left[ {\frac{{{L_S} + {R_S}}}{2}} \right] = \frac{1}{4}E\left[ {{{\left( {{L_S} + {R_S}} \right)}^2}} \right] - {E^2}\left[ {\frac{{{L_S} + {R_S}}}{2}} \right] \nonumber\\
&= {p^2} + \frac{1}{4}\left( { - {{\left( {1 - p} \right)}^n}{p^2} + {p^n}\left( {1 - p} \right)\left( {p + 3} \right) + \frac{{2\left( {1 - \left( {{p^{n + 2}} + {{\left( {1 - p} \right)}^{n + 2}}} \right)} \right)}}{{(n + 1)(n + 2)}}} \right)\nonumber\\
&- {p^2} - p\left( {{p^n}\left( {1 - p} \right) - p{{\left( {1 - p} \right)}^n}} \right) - {\left( {\frac{1}{2}\left( {{p^n}\left( {1 - p} \right) - p{{\left( {1 - p} \right)}^n}} \right)} \right)^2}\nonumber\\
&= \frac{{{p^n}\left( {3{n^2}{{(p - 1)}^2} + 9n{{(p - 1)}^2} + 4(p - 3)p + 6} \right) + p((3n(n + 3) + 4)p + 4){{(1 - p)}^n} - 2{{(1 - p)}^n} + 2}}{{4(n + 1)(n + 2)}}\nonumber\\
&- {\left( {\frac{1}{2}\left( {{p^n}\left( {1 - p} \right) - p{{\left( {1 - p} \right)}^n}} \right)} \right)^2}
\end{align}

The MSE is:

\begin{align}
&MSE\left[ {\frac{{{L_S} + {R_S}}}{2}} \right] = V\left[ {\frac{{{L_S} + {R_S}}}{2}} \right]{\rm{ + bia}}{{\rm{s}}^2}\left[ {\frac{{{L_S} + {R_S}}}{2}} \right] \nonumber\\
&= \frac{{{p^n}\left( {3{n^2}{{(p - 1)}^2} + 9n{{(p - 1)}^2} + 4(p - 3)p + 6} \right) + p((3n(n + 3) + 4)p + 4){{(1 - p)}^n} - 2{{(1 - p)}^n} + 2}}{{4(n + 1)(n + 2)}}
\end{align}

For $L_S$:

\begin{align} \label{E_LS2}
&E\left[ {{L_S}^2} \right] = 0 + n\left( {n - 1} \right)\int_0^p {\int_p^1 {{l^2}{{\left( {l + 1 - r} \right)}^{n - 2}}dr} dl}  + {p^n} \nonumber \\
&= \frac{{p\left( {{n^2}( - p)\left( {{p^n} - 1} \right) - n\left( {p\left( {{p^n} - 3} \right) + 2} \right) - 2(p - 2)\left( {{{(1 - p)}^n} - 1} \right)} \right) - 2{{(1 - p)}^n} + 2}}{{(n + 1)(n + 2)}} + {p^n} \nonumber \\
&= {p^2} - {p^{n + 2}} + {p^n} + \frac{{2\left( {{p^{n + 2}} + n\left( {{p^{n + 2}} - p} \right) - \left( {{p^2} - 2p + 1} \right){{(1 - p)}^n} - 2p + 1} \right)}}{{(n + 1)(n + 2)}}
\end{align}

The variance:

\begin{align}
&V\left( {{L_S}} \right) = E\left[ {{L_S}^2} \right] - {E^2}\left[ {{L_S}} \right] \nonumber\\
&= {p^2} - {p^{n + 2}} + {p^n} + \frac{{2\left( {{p^{n + 2}} + n\left( {{p^{n + 2}} - p} \right) - \left( {{p^2} - 2p + 1} \right){{(1 - p)}^n} - 2p + 1} \right)}}{{(n + 1)(n + 2)}} - \nonumber\\
&{\left( {p + {p^n}\left( {1 - \frac{n}{{n + 1}}p} \right) - \frac{{1 - {{\left( {1 - p} \right)}^{n + 1}}}}{{n + 1}}} \right)^2} \nonumber\\
&= {p^2} - {p^{n + 2}} + {p^n} + \frac{{2\left( {{p^{n + 2}} + n\left( {{p^{n + 2}} - p} \right) - \left( {{p^2} - 2p + 1} \right){{(1 - p)}^n} - 2p + 1} \right)}}{{(n + 1)(n + 2)}} \nonumber\\
&- {p^2} - 2p\left( {{p^n}\left( {1 - \frac{n}{{n + 1}}p} \right) - \frac{{1 - {{\left( {1 - p} \right)}^{n + 1}}}}{{n + 1}}} \right) - {\left( {{p^n}\left( {1 - \frac{n}{{n + 1}}p} \right) - \frac{{1 - {{\left( {1 - p} \right)}^{n + 1}}}}{{n + 1}}} \right)^2} \nonumber\\
&=  - {p^{n + 2}} + {p^n} + \frac{{2\left( {{p^{n + 2}} + n\left( {{p^{n + 2}} - p} \right) - \left( {{p^2} - 2p + 1} \right){{(1 - p)}^n} - 2p + 1} \right)}}{{(n + 1)(n + 2)}} \nonumber\\
&- 2p\left( {{p^n}\left( {1 - \frac{n}{{n + 1}}p} \right) - \frac{{1 - {{\left( {1 - p} \right)}^{n + 1}}}}{{n + 1}}} \right) - {\left( {{p^n}\left( {1 - \frac{n}{{n + 1}}p} \right) - \frac{{1 - {{\left( {1 - p} \right)}^{n + 1}}}}{{n + 1}}} \right)^2}
\end{align}

The MSE:

\begin{align}
&MSE\left( {{L_S}} \right) = V\left( {{L_S}} \right) + bia{s^2}\left( {{L_S}} \right)\nonumber\\
&=  - {p^{n + 2}} + {p^n} + \frac{{2\left( {{p^{n + 2}} + n\left( {{p^{n + 2}} - p} \right) - \left( {{p^2} - 2p + 1} \right){{(1 - p)}^n} - 2p + 1} \right)}}{{(n + 1)(n + 2)}}\nonumber\\
&- 2p\left( {{p^n}\left( {1 - \frac{n}{{n + 1}}p} \right) - \frac{{1 - {{\left( {1 - p} \right)}^{n + 1}}}}{{n + 1}}} \right)\nonumber\\
&= \frac{{{n^2}{{(p - 1)}^2}{p^n} + n(p - 1)\left( {3{p^{n + 1}} - 3{p^n} + 2p{{(1 - p)}^n}} \right) + 2\left( { - 2{p^{n + 1}} + {p^n} + {p^2}{{(1 - p)}^n} - {{(1 - p)}^n} + 1} \right)}}{{(n + 1)(n + 2)}}\nonumber\\
&= \frac{{\left( {(n + 3)n{{(p - 1)}^2} - 4p + 2} \right){p^n} + 2p(n(p - 1) + p){{(1 - p)}^n} - 2{{(1 - p)}^n} + 2}}{{(n + 1)(n + 2)}}
\end{align}

The MSE for $R_S$ is the same, when switching $p$ with $1-p$.

\subsection{Endpoint values for the parameter $p$}

It has been assumed throughout that there are truly two classes to differentiate between. In principle, $p$ could be allowed to be in $[0,1]$ instead of $(0,1)$, to include the possibility - unknown to the user - that one of the two classes is empty. The analysis for endpoint $p$ is different, because the central case where $0<L<R<1$, that has probability approaching $1$ as $n$ increases (if $p$ is strictly between $0$ and $1$), has probability identically zero under $p=0$ or $p=1$. Details are omitted. The case where $p$ is $0$ or $1$ is not too relevant for the practical purpose of decision trees.


\subsection{Proof that $(L, R)$ is a \textit{minimal non-complete} sufficient statistic (for n=2)}
\label{app:LR_min_suff}

Let there be two sets $(\underline{X}_1, \underline{Y}_1)$ and $(\underline{X}_2, \underline{Y}_2)$ where both have the same sufficient statistic (i.e.: $T(\underline{X}_1, \underline{Y}_1) = T(\underline{X}_2, \underline{Y}_2) = (L, R)$). Then it is clear that $ \frac{f(\underline{X}_1, \underline{Y}_1)}{f(\underline{X}_2, \underline{Y}_2)} = 1$ (as long as the ratio is defined) is not dependent on p. Since the opposite is also evident then $(L, R)$ is a \textit{minimal} sufficient statistic. That is, the pair $(L, R)$ most efficiently captures all possible information about the unkown parameter $p$ from the supervised uniform distribution.

In order to prove non-completeness, a non-zero function $g$ should be found such that $E[g(T)]=0$ regardless of the value of $p$. Consider for $n=2$ the function
\begin{align} 
	g\left( {L, R} \right) = \left\{ {\begin{array}{*{20}{c}}
			{2 - \frac{1}{L}}&{0 < L<R = 1} \\ 
			2&{0 < L<R < 1} \\ 
			{2 - \frac{1}{{1 - R}}}&{0 = L<R < 1} 
		\end{array}} \right.
		\label{completness_01}
	\end{align}
	It is immediate to check using eq. (\ref{E_g_LR}) that $g$ is a non-constant unbiased estimator of $0$: leaving aside the constant $2$, observe that the integrands in the first and third summands in eq. (\ref{completness_01}) are $1$ so their integral is $2(1-p)+2p=2$.  Hence, $(L, R)$ is not complete for $n=2$. 
	
	

\subsection{Extra figures}
\label{app:extra_figures}

\subsubsection{Performance of estimators for $p$}
\label{sec:performance_of_p_est}

\begin{figure}[h]
	\centering
	\includegraphics[width=0.99\linewidth]{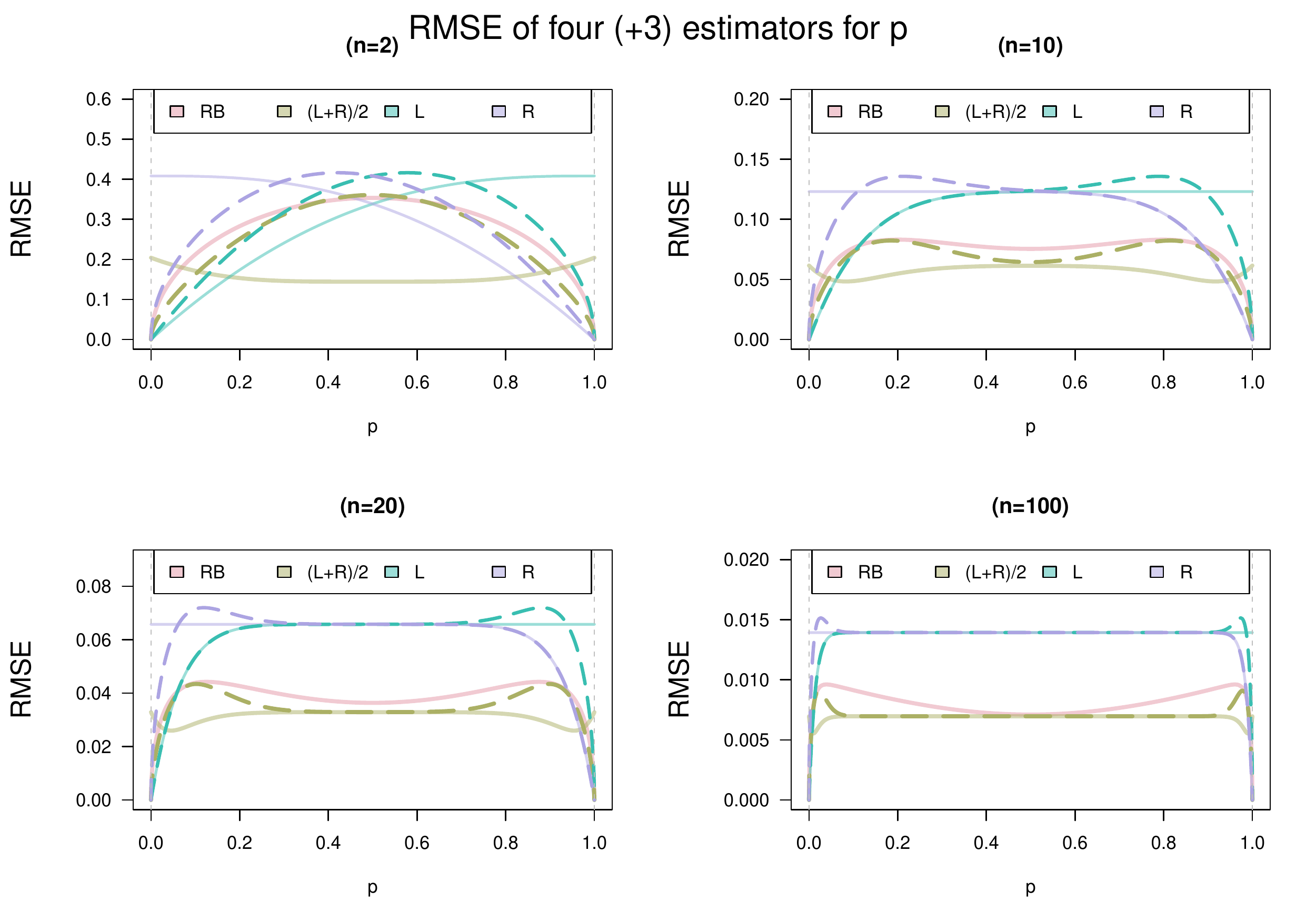}
	\caption{The RMSE of the four estimators + the three sweep estimators (in dashed lines), across $p$, for different sample sizes ($n=2,10,20,100$).}
	\label{fig:RMSE_curves_sweep}
\end{figure}

\subsubsection{Simulations}

\begin{figure}[h]
	\centering
	\includegraphics[width=0.99\linewidth]{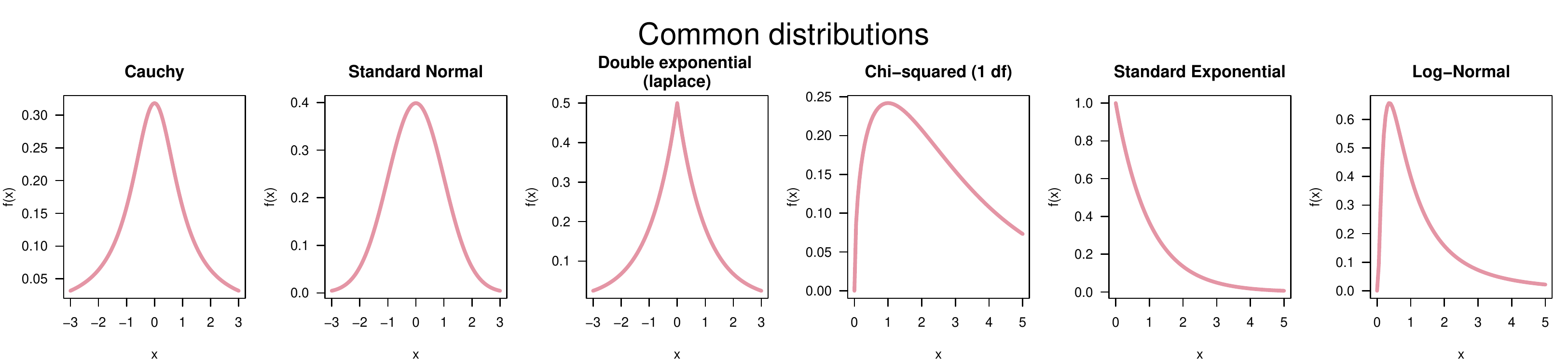}
	\caption{The densities for six well known distributions: Cauchy, Standard Normal, Double exponential (laplace), Chi-squared (1 df), Standard Exponential, and Log-Normal}
	\label{fig:common_dist_densities}
\end{figure}

\begin{figure}[h]
	\centering
	\includegraphics[width=0.99\linewidth]{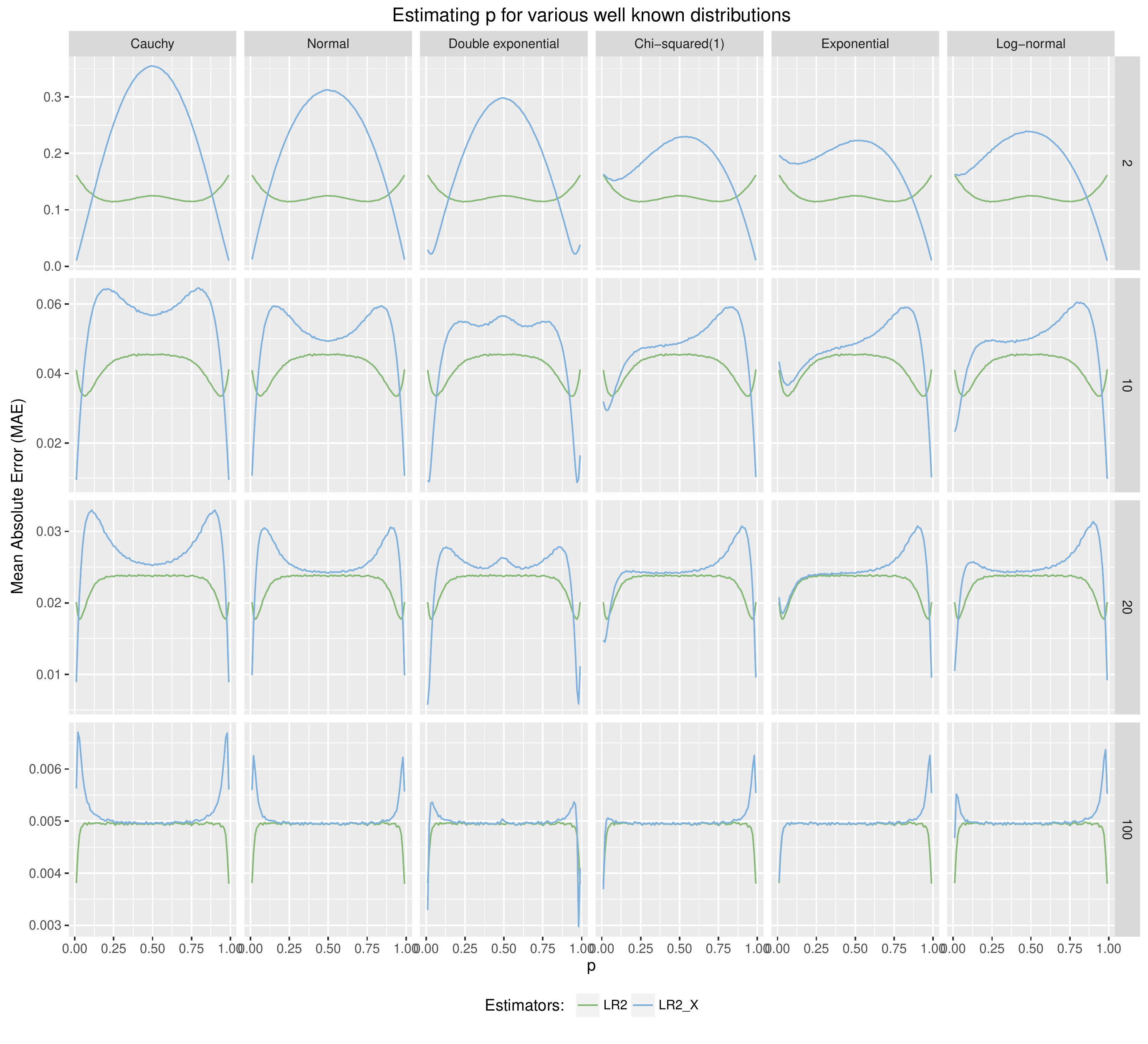}
	\caption{The misclassification error (Mean Absolute Error) of a model estimating $p$ using the original data or the quantile transformed data, for a range of $p$ positions, sample sizes, for six well known distributions: Cauchy, Standard Normal, Double exponential (laplace), Chi-squared (1 df), Standard Exponential, and Log-Normal. In the figure $LR2=\hat p_{B}$ and $LR2_X=\hat p_X$.}
	\label{fig:Common_dist_simu_LR2_LR2X}
\end{figure}

The same simulation was performed for six well known distributions: Cauchy, Standard Normal, Double exponential (laplace), Chi-squared (1 df), Standard Exponential, and Log-Normal. Figure \ref{fig:common_dist_densities} presents their densities, and Figure \ref{fig:Common_dist_simu_LR2_LR2X} compares the performance of using $\hat p_B = LR2= \frac{L+R}{2}$ vs $\hat p_X = LR2_X= \frac{L_X+R_X}{2}$. The former could be calculated only if the distribution's CDF is known so that the quantile transformation could be used. 

In unimodal and right-tailed distributions (Chi-squared (1 df), Standard Exponential, and Log-Normal) the asymmetry is reflected in that the performance of both estimators becomes nearly the same for values of $p$ nearing 0, while $\hat p_B$ beats $\hat p_X$ for values of $p$ nearing 1.


%

\subsection{Using the ECDF for estimating the distribution of $X$}

The method which relies on the smallest number of assumptions is to estimate the CDF using the step-function empirical distribution function. This is realistic when there is a lot of unlabeled observations that could be used to estimate the distribution - this assumes that both the training, testing and unlabeled observations are all i.i.d and comes from the same distribution. This is often called semi-supervised learning, and is feasible in scenarios were unlabeled data is cheap but labeled data is expensive.


To test this strategy, a simulation was done to see the MAE of when using the ECDF. The distribution of the predictor variable is standard normal. The simulation tested the performance of the parameter estimation (in terms of MAE) for different ratios of unlabeled to labeled observations. A ratio of 0 means only the labeled data is used. This is similar to using the rank transform on the observations of the training set. A ratio of $\frac{3}{2}$ is what we would expect to get if all the labeled data was transformed using its own CDF, but then that the misclassification error would be tested using a bootstrap sample (as is used in bagging or random forest models), since the bootstrap often uses only two thirds of the data for training and is left with another third for testing (and also for the ECDF estimation/transformation). Higher ratios of 1, 5, 10 and 100 are similar to what one could expect in various real world datasets (as do also sometimes appear in data science competitions such as the ones placed on Kaggle).

\begin{figure}
	\centering
	\includegraphics[width=0.99\linewidth]{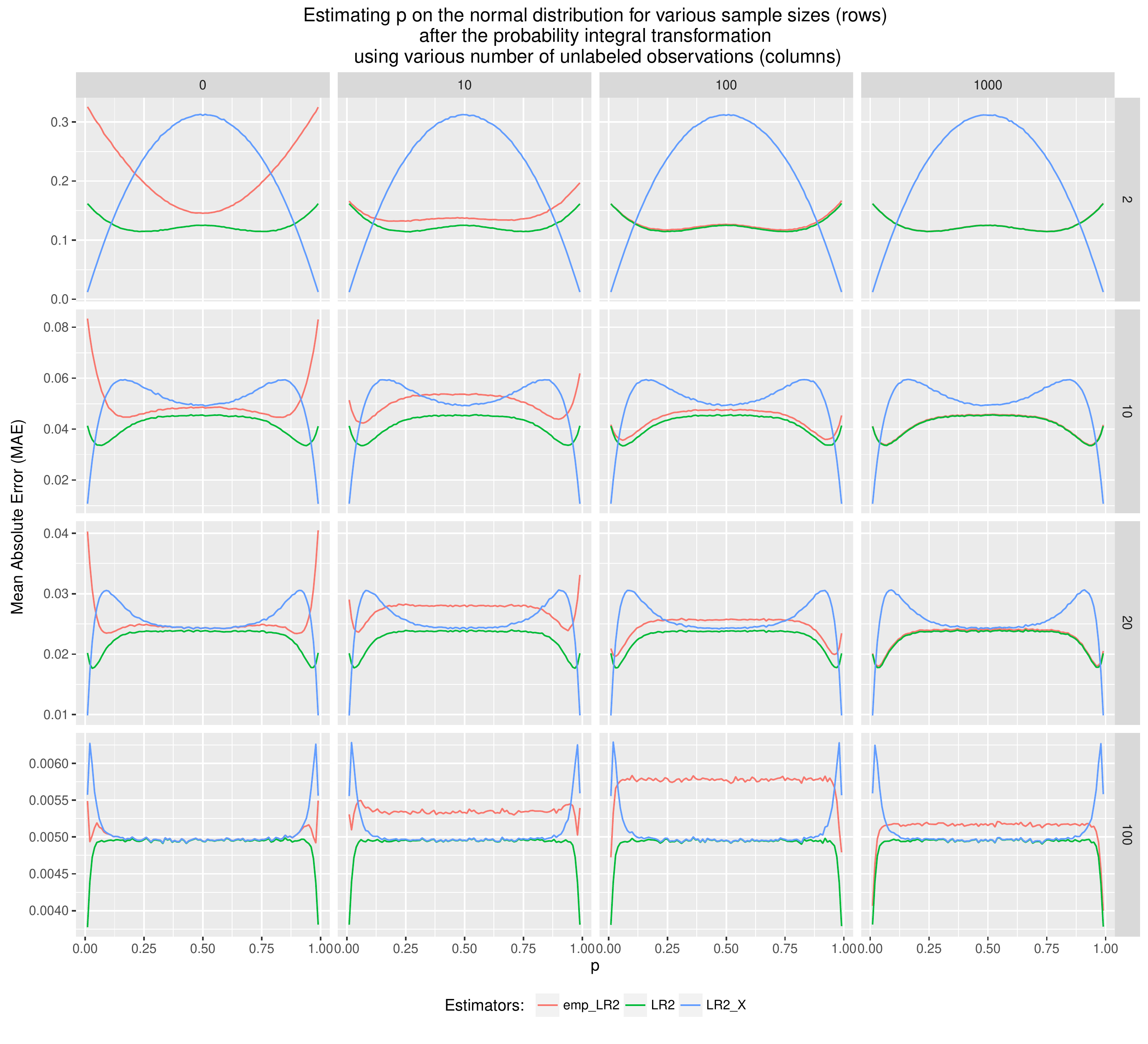}
	\caption{The misclassification error (Mean Absolute Error) of a model estimating $p$ using the original data which comes from a standard normal distribution (blue), estimation after the quantile transformed data (green), and after using the empirical quantile transformation (red) - for a range of $p$ positions, sample sizes. The unlabeled sample sizes (columns) are 0, 10, 100 and 1000. In the figure $LR2=\hat p_{B}$ and $LR2_X=\hat p_X$.}
	\label{fig:ECDF_semisupervised_simu_LR2_LR2X_norm}
\end{figure}

There are n observations sampled from the the standard normal and standard exponential distributions. Each row in the graph is for a different number of observations (2, 10 20, and 100). Each columns is the number of unlabeled observations that are used to estimate the ECDF. 
The blue line is estimating $p$ using ${(Lx+Rx)} \over 2$ (we actually use F on the result to get it to the $U(0,1)$ distribution so that we could calculate its MAE from $p$).
The green line is estimating $p$ using ${(L+R)} \over 2$ (as if we knew the real CDF).
The red line is estimating $p$ using ${(eF(Lx)+eF(Rx))} \over 2$ - where eF is the empirical CDF estimated by using both the labeled and unlabeled data together.
For example, the first row and the second column from the left is for when there are 2 labeled observations and 10 unlabeled observations (a total of 12 observations for estimating the CDF). As can be seen in both Figure \ref{fig:ECDF_semisupervised_simu_LR2_LR2X_norm} and Figure \ref{fig:ECDF_semisupervised_simu_LR2_LR2X_exp} (in the appendix), the red line is doing quite well (very close to the green line).

\begin{figure}
	\centering
	\includegraphics[width=0.99\linewidth]{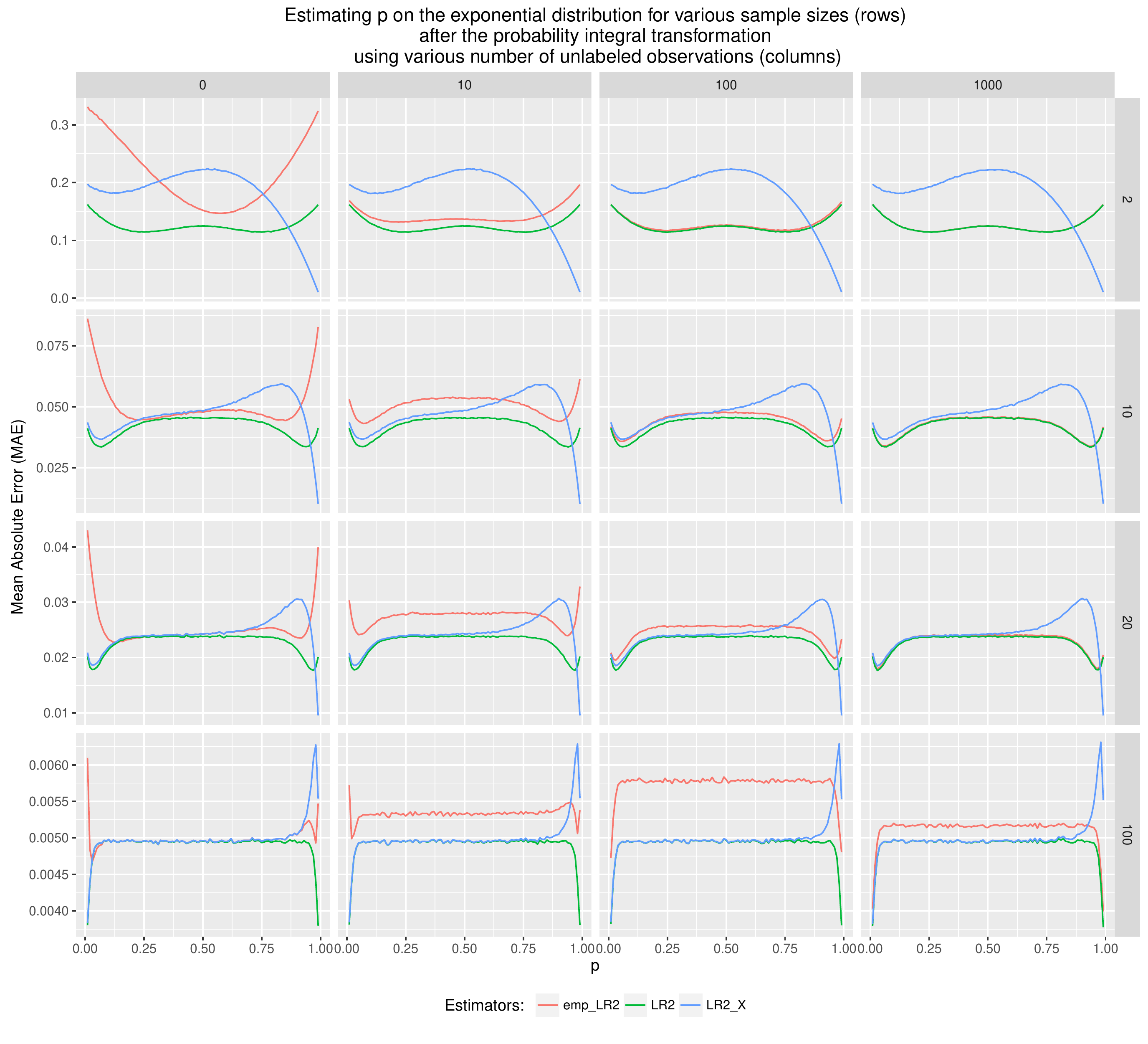}
	\caption{The misclassification error (Mean Absolute Error) of a model estimating $p$ using the original data which comes from an exponential distribution (blue), estimation after the quantile transformed data (green), and after using the empirical quantile transformation (red) - for a range of $p$ positions, sample sizes. The unlabeled sample sizes (columns) are 0, 10, 100 and 1000.}
	\label{fig:ECDF_semisupervised_simu_LR2_LR2X_exp}
\end{figure}

The results from both Figure \ref{fig:ECDF_semisupervised_simu_LR2_LR2X_norm} and Figure \ref{fig:ECDF_semisupervised_simu_LR2_LR2X_exp} show that it requires about an order of magnitude more unlabeled samples in order to get the needed precision. For 2 labeled data, the results are reasonable if there are 10 unlabeled observations. And for 10 labeled data, the results are reasonable if there are 100 unlabeled observations - the estimation of the CDF is good enough so that the empirical transformation yields almost the same results as the using the CDF of the real distribution (normal). But 100 unlabeled observations is already not enough for 20 labeled data, while 1000 unlabeled data will be enough. And for 100 labeled data, a 1000 unlabeled data is not enough. Hence, if there are not enough unlabeled observations, it could arguably not be worth using the unlabeled ones since the error produced by the CDF estimation is greater than the error produced by not using the real CDF but the original data instead. Also, using the ECDF of the labeled observations on themselves (left most column) shows how the transformations biases the estimation of $p$ so that it would favor value of $p$ nearing 0.5, so by itself it is not recommended. 


%
%


\subsection{Models requiring more than one split} \label{sec:more_than_one_split}


The single split point estimation is the simplest model a decision rule needs to estimate, and is in fact what the decision tree needs to do at every node of the tree. Until now, this paper focused on the number of observations needed for estimating a single split, and the way that knowing the underlaying distribution can help to use the optimal Bayes estimator (under a uniform prior). However, the previous results may be misleading in that many real-life datasets require the estimation of very complex models, such that each split is practically left with a rather small sample size.

This section explores how introducing more than one split can easily strain the decision tree prediction accuracy, for a dataset that is perfectly explained by the variables in a series of rules which are horizontal or vertical to the axes. All simulations will focus on $p=0.5$ (that is, the chances of an observation getting 1 or 0 is 0.5 each), and the distribution of the original observations will be normal. The observations will such that for every $i=1,...n$, $X_i \sim N(0,1)$ and $U_i = F(X_i) \sim U(0,1)$. 

\begin{figure}[h]
	\centering
	\includegraphics[width=0.8\linewidth]{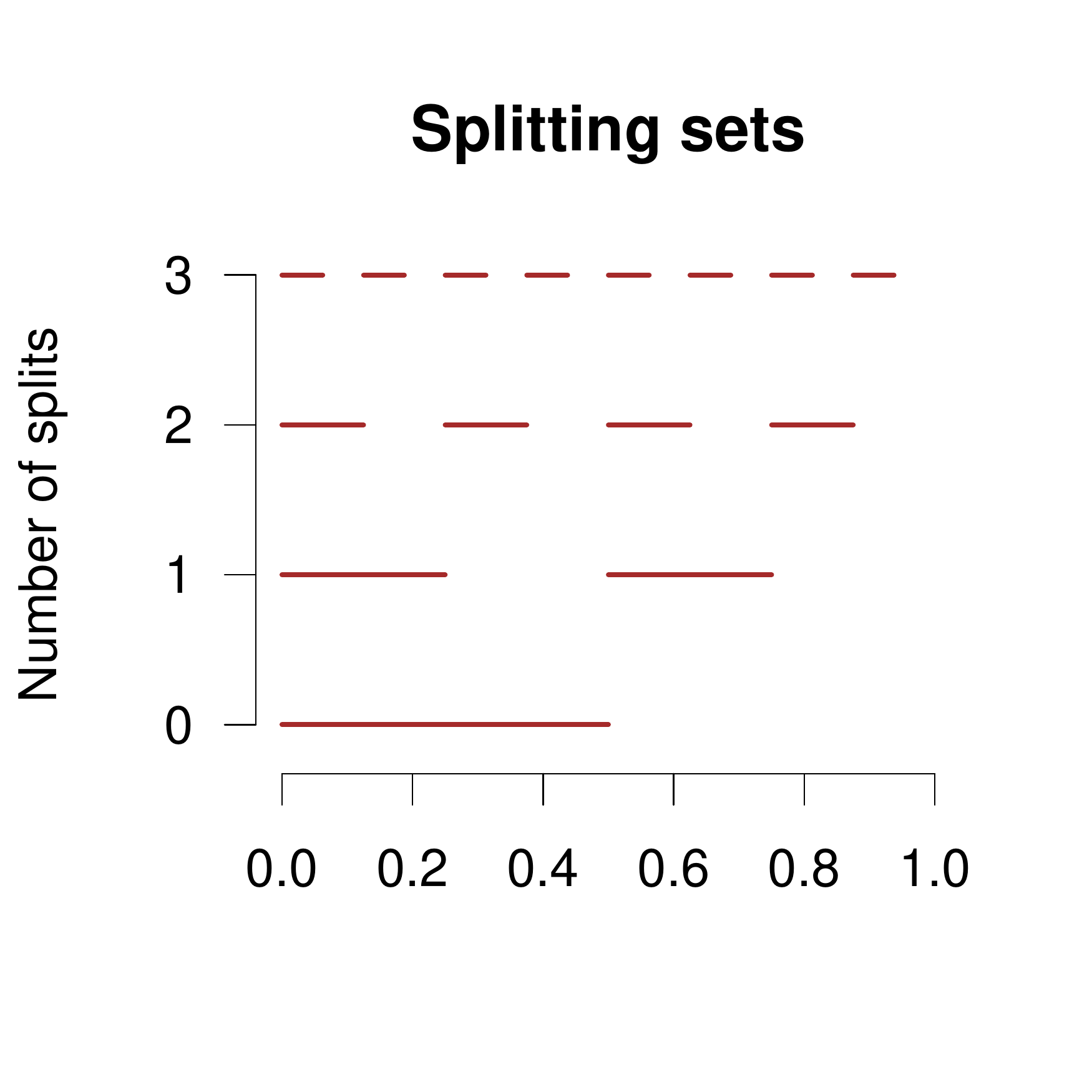}
	\caption{A series of sets, for each one the chances of getting 1 is 0.5, but the number of splits is different.}
	\label{fig:splitting_sets_example}
\end{figure}

The "splitting sets" series will be defined in the region of 0 to 1. A splitting set with 0 splits will be one where if $U_i<0.5$ and zero otherwise. A splitting set with 1 split will get one if $U_i<0.25$ or $0.5<U_i<0.75$, and zero otherwise. A splitting set with 2 split will get one if $U_i<0.125$ or $0.25<U_i<0.375$ or $0.5<U_i<0.625$ or $0.75<U_i<0.875$, and zero otherwise. Figure \ref{fig:splitting_sets_example} presents the splitting sets for different orders (0 to 4), each set defines rules that will assure that the chances of getting 1 is 0.5. The case of a split set with 0 splits is what was shown until now when $p=0.5$.

As was shown in Figure \ref{fig:normmix_dist_simu_LR2_LR2X}, the difference between $\hat p_B$ and $\hat p_X$ gets smaller for $p=0.5$ in the normal distribution, as the sample size increases. 

\begin{figure}[h]
	\centering
	\includegraphics[width=0.8\linewidth]{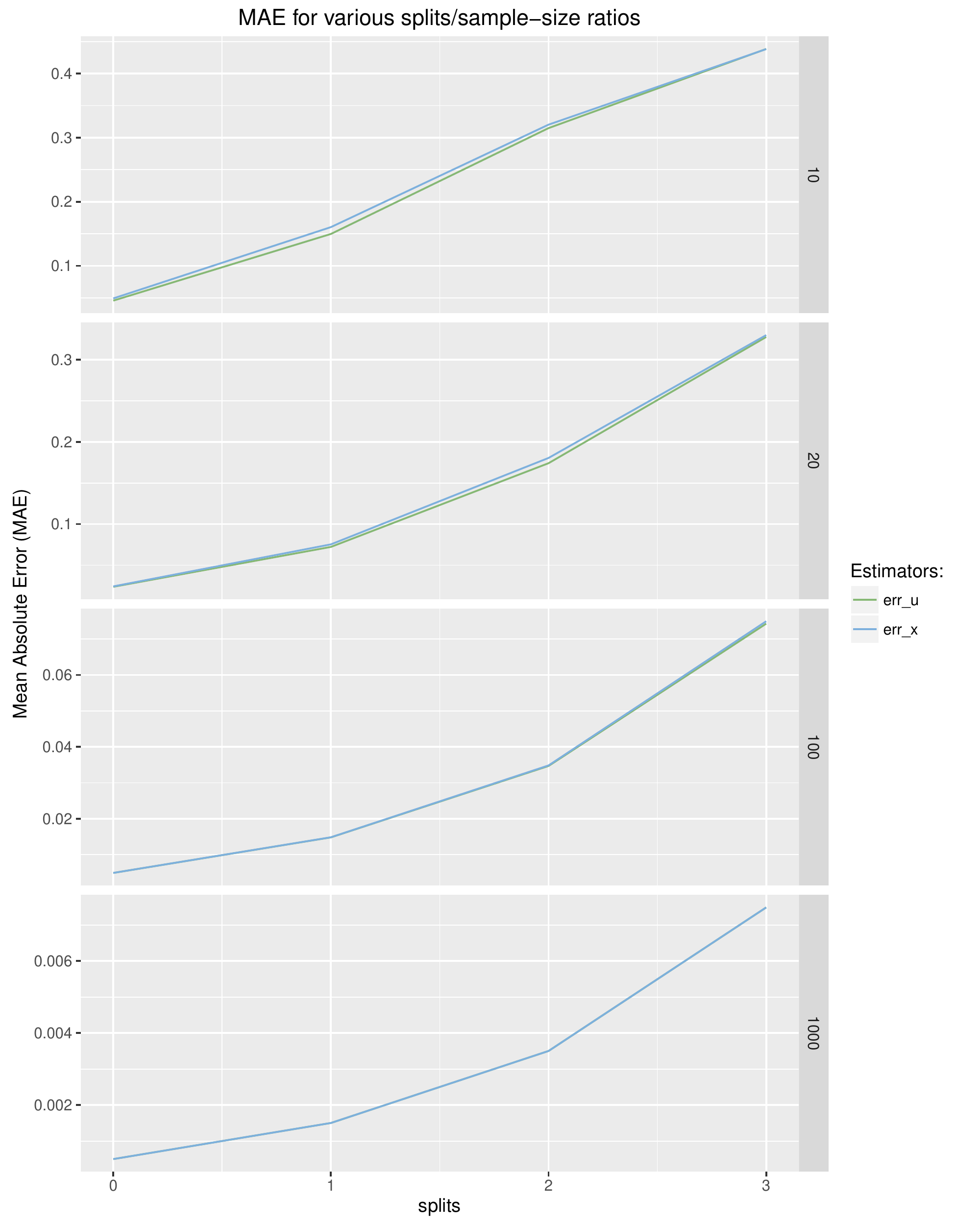}
	\caption{The misclassification error (MAE) in various sample sizes (10, 20 and 100), for comparing $\hat p_B$ and $\hat p_X$ on a split-set with different number of splits (0 to 3). Each data point is based on $10^5$ simulations}
	\label{fig:split_sets_1d_MAE_per_n}
\end{figure}

\begin{figure}[h]
	\centering
	\includegraphics[width=0.8\linewidth]{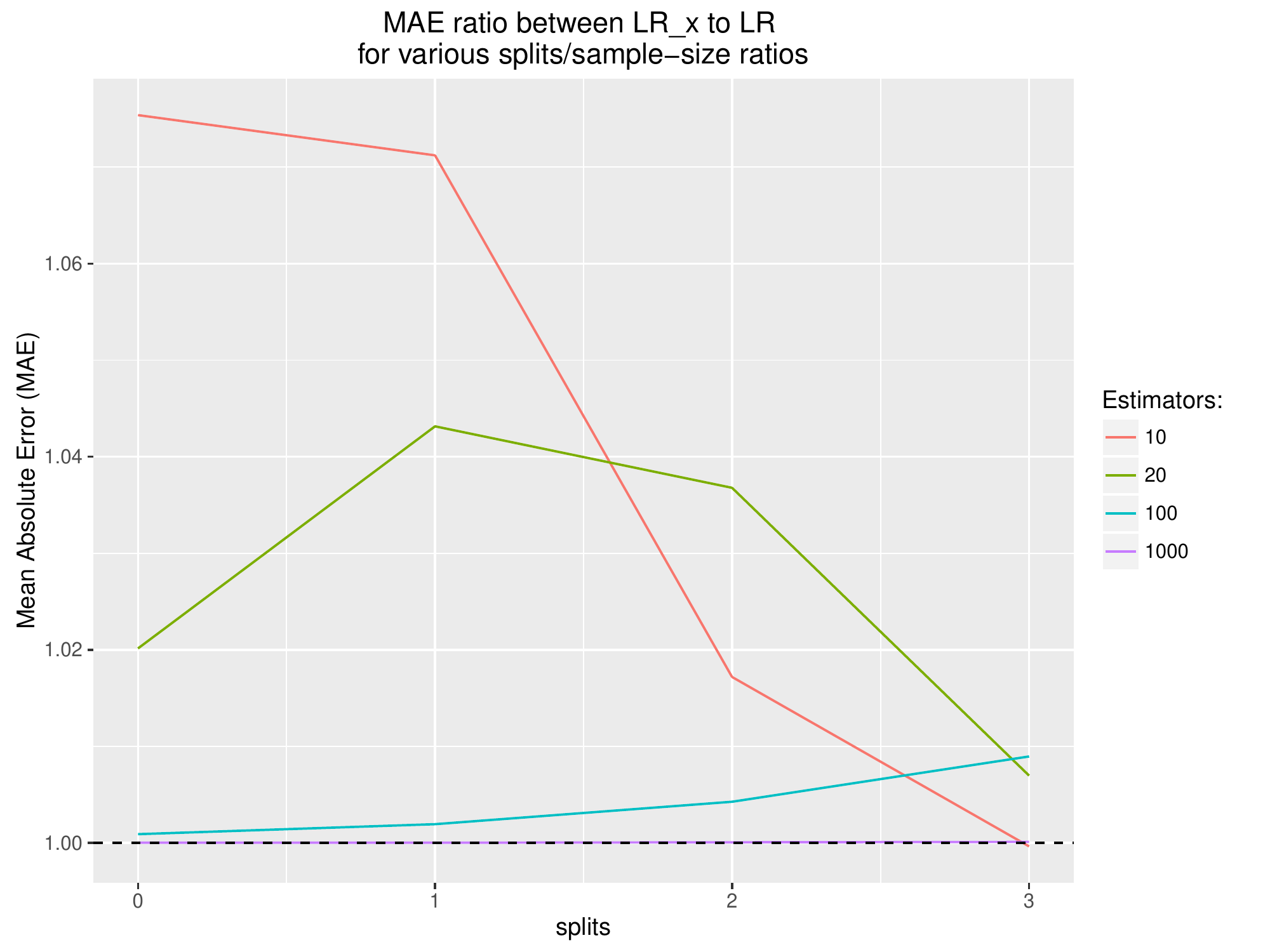}
	\caption{The ratio in misclassification error (MAE) between $\hat p_X$ and $\hat p_B$, in various sample sizes (10, 20 and 100), on a split-set with different number of splits (0 to 3). Each data point is based on $10^5$ simulations}
	\label{fig:split_sets_1d_MAE_ratio_per_n}
\end{figure}

It is seen on Figure \ref{fig:split_sets_1d_MAE_per_n} that the more splits are in the split set, the higher the MAE. It is also seen that the advantage of using $\hat p_B$ over $\hat p_X$ for the different configurations. The exact order of improvement depends on the order of splits and sample size. For 1000 observations there is practically no difference between using the transformation or not. For only 10 observations the value of the transformation decreases the larger the number of splits, this is probably because with 3 splits, it is not enough to have just 10 observations. In general, it seems that there is a complex relationship between the complexity of the required model (order of number of splits) and the number of available observations. See Figure \ref{fig:split_sets_1d_MAE_ratio_per_n} for the results.


\begin{figure}[h]
	\centering
	\includegraphics[width=0.8\linewidth]{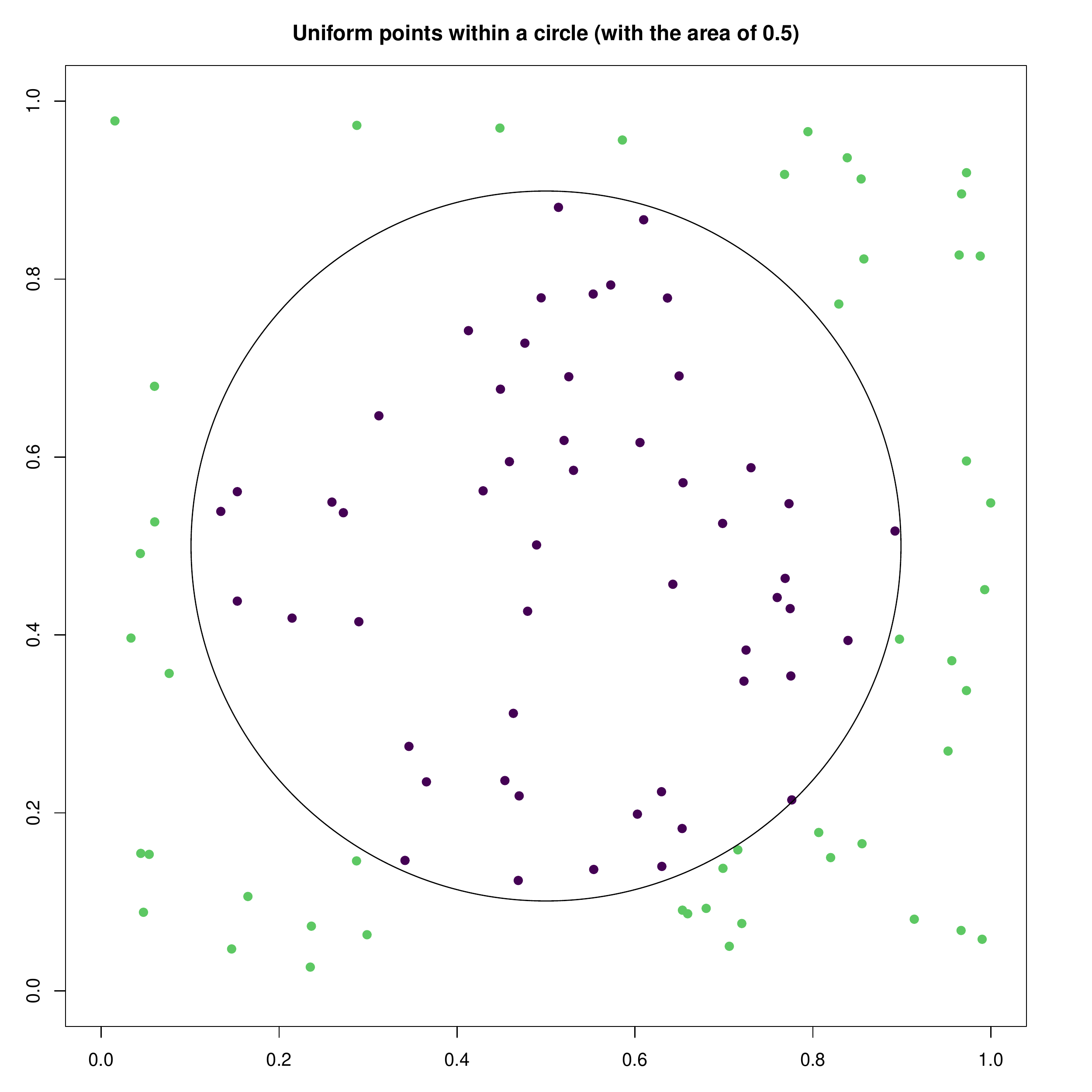}
	\caption{A circle with the area of 0.5 inside the unite square, when sampling 100 observations}
	\label{fig:splitting_circle_2d_plot}
\end{figure}

In the following example, a circle with the area of 0.5 is present in the center of the unit square. Observations are sampled uniformly within this square and labeled if they are inside or outside of the circle, see Figure \ref{fig:splitting_circle_2d_plot}.

\begin{figure}[h]
	\centering
	\includegraphics[width=0.99\linewidth]{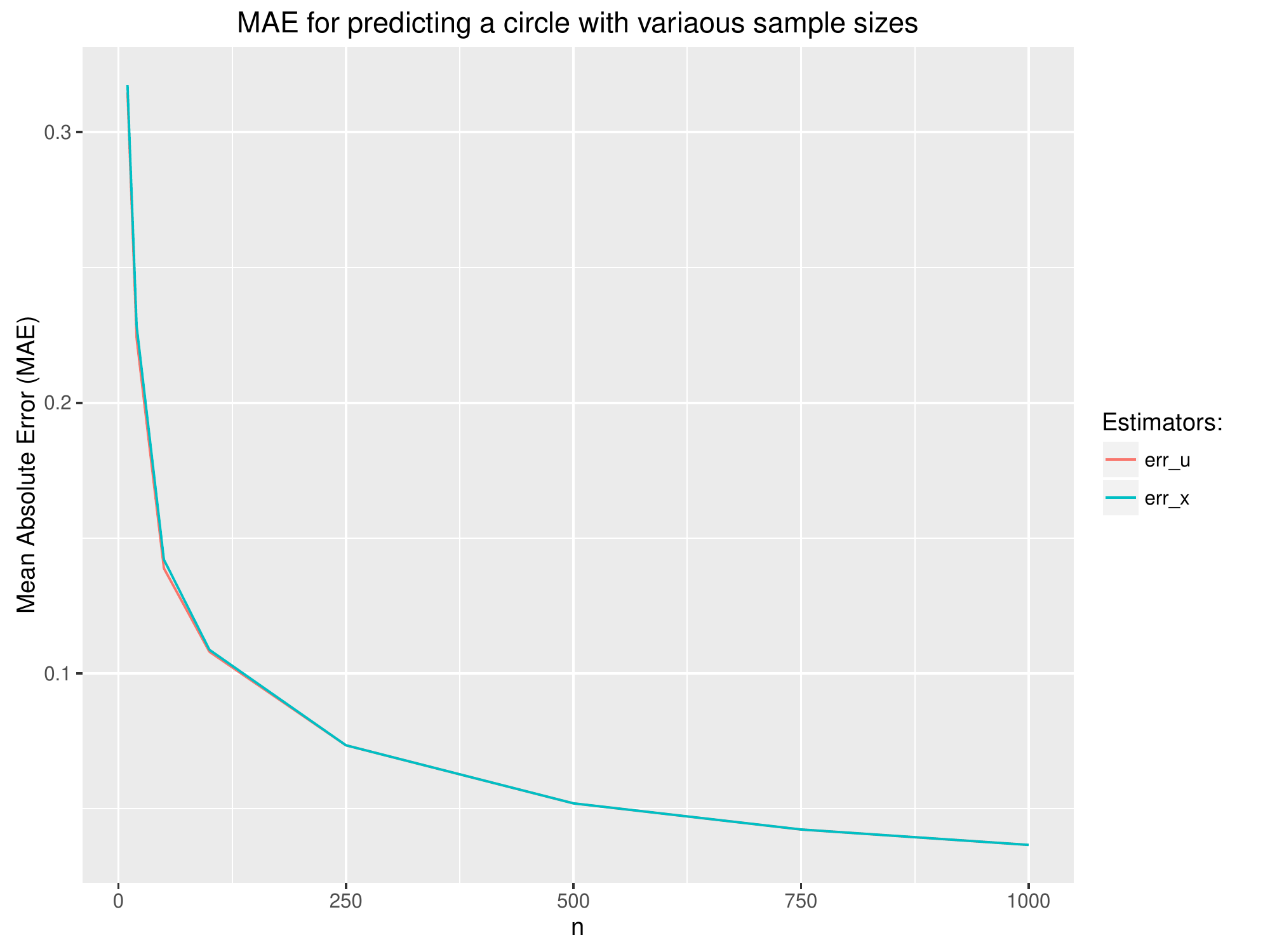}
	\caption{The misclassification error (MAE) in various sample sizes (10,20, 50, 100, 250,500,750, and 1000), for comparing $\hat p_B$ and $\hat p_X$ on predicting the region of a circle, using the quantile scale and when using the normal scale. Each data point is based on $10^5$ simulations }
	\label{fig:split_sets_circle_2d_MAE_per_n}
\end{figure}

\begin{figure}[h]
	\centering
	\includegraphics[width=0.8\linewidth]{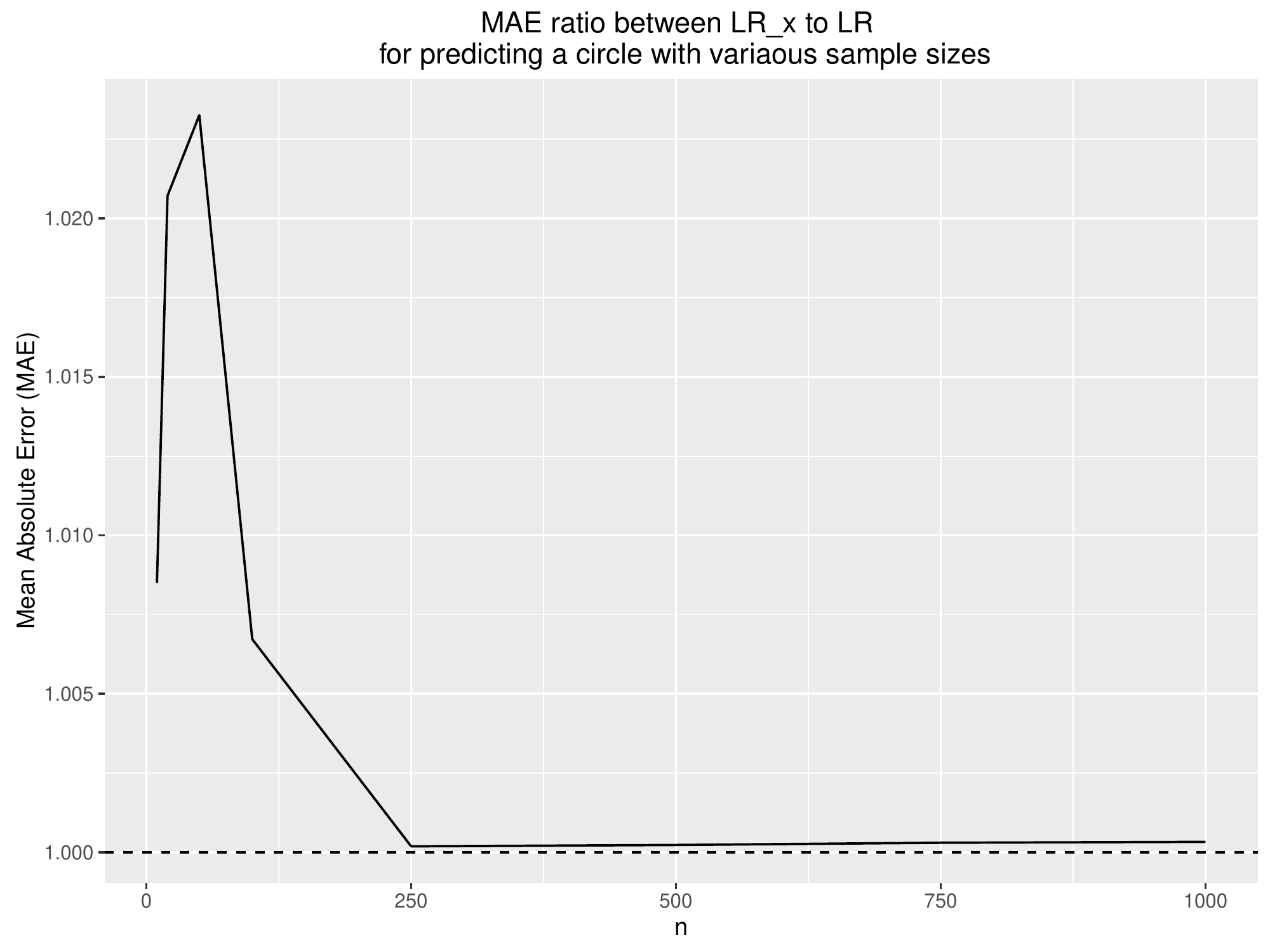}
	\caption{The ratio in misclassification error (MAE) between $\hat p_X$ and $\hat p_B$, in various sample sizes (10,20, 50, 100, 250,500,750, and 1000), for comparing $\hat p_B$ and $\hat p_X$ on predicting the region of a circle, using the quantile scale and when using the normal scale. Each data point is based on $10^5$ simulations }
	\label{fig:split_sets_circle_2d_MAE_ratio_per_n}
\end{figure}

With this labeled data, a decision tree is trained in order to estimate the regions inside and outside of the circle. The model is fitted using various sample sizes: 10, 20, 50, 100, 250, 500, 750, 1000. And for each of them, the model is then tested on 160,000 points in a grid within the square to evaluate the misclassification error of the model. This process is repeated 100,000 times, and the mean absolute error is numerically derived. These are done once when the two dimensional predictor space is uniform and also when it is standard normal. Figure \ref{fig:split_sets_circle_2d_MAE_per_n} shows the MAE for various sample sizes, and it is clear that using the quantile scale always gives better accuracy, but just slightly.  Figure \ref{fig:split_sets_circle_2d_MAE_ratio_per_n} helps to highlight the benefit by comparing the ratio of MAE between the methods. The interesting pattern that emerges is that for large sample size of 1000, there is no benefit to the method, but that for medium sample sizes (from 10 to 100 observations), the benefit can be between 1.005 to 1.024 times improvement in the MAE (when transforming the observations to the uniform distribution).

\subsection{When the data for training has a different distribution than the data for testing}

So far this work assumed that the data used for training the decision tree model comes from the same distribution as the data used for testing the model. There are cases when this is not the case. For example, in case-control studies, usually the proportion of "cases" in the training set is larger than their proportion in the population. When this occurs, in order to train the decision tree for optimizing misclassification error on the test data (assume equal cost for misclassifying cases or controls), it is possible to add pre-specified prior probabilities to the case control based on prior knowledge on their proportion in the population, where the model would eventually be used. In the case of split-point interpolation, this strategy is good as long as the only difference between the train and test datasets is only the proportion of cases and controls, but that the distribution of observations in the predictor variable is the same as the population. Two counter examples are presented in this section for demonstrating how different distributions in the predictor variables can have devastating results on prediction, even when the class probabilities are the same between train and test.

\subsubsection{Example 1: When the predictor variable in train has only two value while test is continuous}

Let an observation from the population (test data) be the couple $\left<X_{test}, Y_{test}\right> \sim SU(0,1, p)$ so that the predictor variable $X_{test}$ comes from $U(0,1)$ and the response variable $Y_{test}$ will get 1 if $X_{test}<p$ and 0 otherwise. However, an observation from the train dataset $\left<X_{train}, Y_{train}\right>$ is one where $X_{train} \sim B(1, p)$ and $Y_{train}$ will get 1 if $X_{train}<p$ and 0 otherwise. If a DTL will be trained on $\left<X_{train}, Y_{train}\right>$ and then its prediction accuracy tested on $\left<X_{test}, Y_{test}\right>$ there would be no reason to adjust the prior distributions of $\left<X_{train}, Y_{train}\right>$ since $E[Y_{train}]=E[Y_{test}]=p$. Since $X_{train}$ can have only two values, the split-point interpolation will be fixed no matter what is $p$. If using $L=0$ then the misclassification error (MAE) would always be $p$, and if it uses $R=1$ then it would be $1-p$. If the interpolation is done using $\frac{L+R}{2}=\frac{1}{2}$ then the MAE would be $|\frac{1}{2}-p|$.

\subsubsection{Example 2: When the predictor variable in both train and test are continuous but from different distributions}

Again, let an observation from the population (test data) be the couple $\left<X_{test}, Y_{test}\right> \sim SU(0,1, p)$ so that the predictor variable $X_{test}$ comes from $U(0,1)$ and the response variable $Y_{test}$ will get 1 if $X_{test}<p$ and 0 otherwise. 
	
The observations from $\left<X_{train}, Y_{train}\right>$ are such where $X_{train} \sim BT(p)$ and $Y_{train}$ will get 1 if $X_{train}<p$ and 0 otherwise.

\begin{figure}[h]
	\centering
	\includegraphics[width=0.5\linewidth]{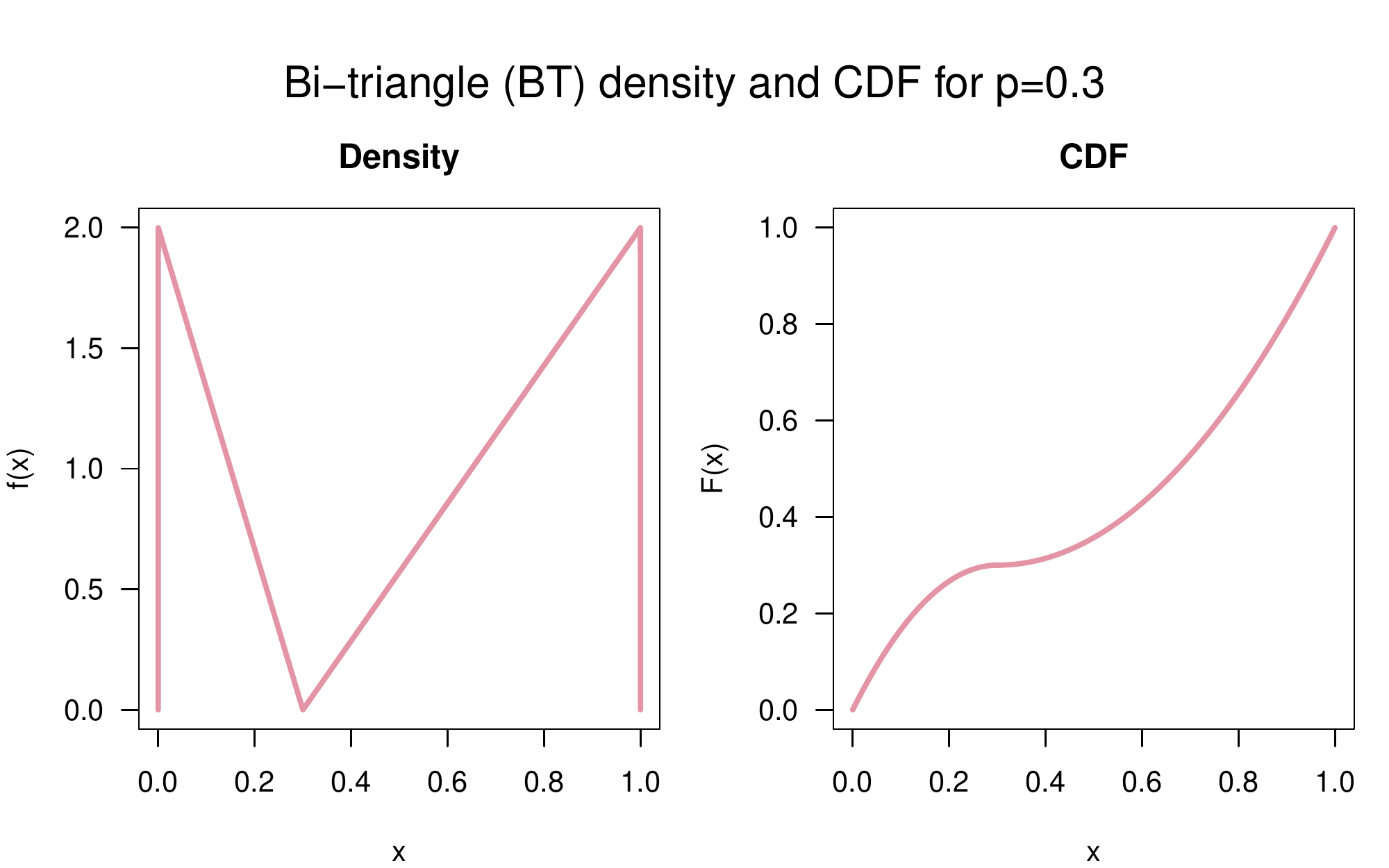}
	\caption{The density and CDF for the Bi-triangle distribution with $p=0.3$.}
	\label{fig:bitriangle_density_CDF}
\end{figure}

A random variable $X$ is from the Bi-triangle distribution (BT) with the parameter $p$ and support $[0,1]$ if it has the following density and CDF (see Figure \ref{fig:bitriangle_density_CDF} for the distribution for $p=0.3$):

\begin{align}
{f_X}\left( x \right) &= {I_{x < p}}\left( {2 - \frac{2}{p}x} \right) + {I_{x > p}}\left( { - \frac{2}{{1 - p}}p + \frac{2}{{1 - p}}x} \right) \\
{F_X}\left( x \right) &= p - {I_{x < p}}\frac{{\left( {p - x} \right)}}{2}\left( {2 - \frac{2}{p}x} \right) + {I_{x > p}}\frac{{\left( {p - x} \right)}}{2}\left( { - \frac{2}{{1 - p}}p + \frac{2}{{1 - p}}x} \right)
\end{align}

\begin{figure}[h]
	\centering
	\includegraphics[width=0.99\linewidth]{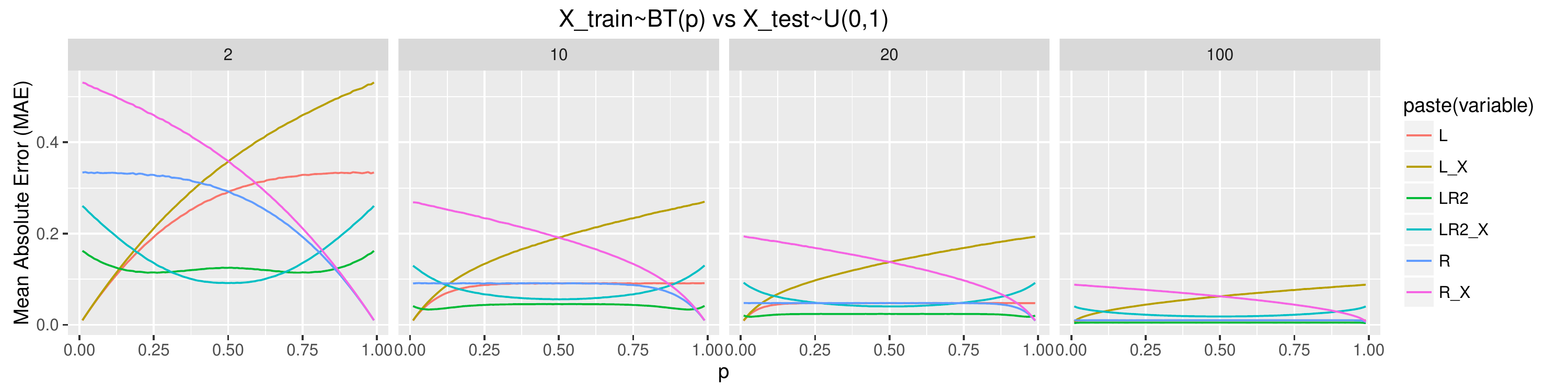}
	\caption{Misclassification error when predicting test data when $X_{test} \sim U(0,1)$ while $X_{train} \sim BT(p)$. In the figure $LR2=\hat p_{B}$ and $LR2_X=\hat p_X$.}
	\label{fig:bitriangle_simu_p_est}
\end{figure}

If a DTL will be trained on $\left<X_{train}, Y_{train}\right>$ and then its prediction accuracy tested on $\left<X_{test}, Y_{test}\right>$, as before, there would be no reason to adjust the prior distributions of $\left<X_{train}, Y_{train}\right>$ since $E[Y_{train}]=E[Y_{test}]=p$. Figure \ref{fig:bitriangle_simu_p_est} presents the misclassification errors of predictions. In the previous sections the performance of using $\frac{L_X+R_X}{2}$ often had a lower MAE than using $L$ or $R$, and no more than twice $\frac{L+R}{2}$. However, this example shows how the MAE is more than twice, and how using $L_X$ gives a much higher MAE than $L$ (as opposed to when the distribution of $X_{train}$ was the same as $X_{test}$).

If the distributions of both $X_{train} \sim F_{train}$ and $X_{test} \sim F_{test}$ are known (or at least well approximated using unlabeled data), this problem could be resolved through the quantile transformation by simply fitting the model on to $F_{train}(X_{train}) \sim U(0,1)$ and then predicting new observations using $F_{test}(X_{test}) \sim U(0,1)$.

\subsection{Reproducible research}
\label{app:reproducible}


The R code for reproducing the simulations:

\begin{itemize}

	\item Figures \ref{fig:RMSE_curves} 
	\ref{fig:MAE_curves}, and \ref{fig:RMSE_curves_sweep} are used to illustrated the comparison of different estimators in the supervised uniform setting. The lines have been analytically calculated except for the MAE of the Rao-Blackwell and some swept estimators, which were estimated through simulation ($10^5$ times for every point). These can be reproduced based on the code in the files \textit{RMSE\_curves.R}, \textit{MAE\_curves.R}, \textit{RMSE\_curves\_Sweep.R}.
	
	\item Figures \ref{fig:Beta_dist_densities}
	,\ref{fig:Beta_simu_LR2_LR2X}, 	\ref{fig:normmix_dist_simu_LR2_LR2X}, and 
	\ref{fig:mixnorm_dist_densities} - all deal with the densities and simulations on the Beta and mixture of normal distributions. These can be reproduced based on the code in the files \textit{beta\_simulations.R} and \textit{bi\_normal\_simulations.R}.
	
	\item Figure \ref{fig:ECDF_semisupervised_simu_LR2_LR2X_norm} - the simulation of using the ECDF to correct X for split point interpolation - can be reproduced based on the code in the file \textit{UL\_semiparametric\_MAE\_simulations.R}. This simulation can take up to a day to complete. The simulation is run $10^5$ times for every point.
	The simulation for Figure  \ref{fig:norm_musigma_est_simu_LR2_LR2X_norm} can be reproduced using the code in \newline	\textit{UL\_parametric\_estimation\_MAE\_simulations.R}.
	
	\item Table \ref{table:weather1} - the weatherAUS example - can be reproduced based on the code and output in the files: \textit{example\_weatherAUS.Rmd} and \textit{example\_weatherAUS.html}.
	
	\item split sets simulations are organized in \textit{split\_sets\_many\_splits\_to\_n.R}
\end{itemize}

In the process of this work we wrote the edfun R package \citep{R_edfun} - a package for easily creating empirical distribution functions from data: 'dfun', 'pfun', 'qfun' and 'rfun' - that run fast for the purpose of simulation. This package was used in order to create the needed CDF and inv-CDF functions for the double exponential distribution and the mixture model of two normal distributions.

%
%
%
%
%
%
%
%
%

\end{document}